\documentclass[10pt]{article} 
\usepackage[accepted]{tmlr}

\usepackage{amsmath,amsfonts,bm}

\def\eqref#1{equation~\ref{#1}}

\def\1{\bm{1}}

\DeclareMathAlphabet{\mathsfit}{\encodingdefault}{\sfdefault}{m}{sl}
\SetMathAlphabet{\mathsfit}{bold}{\encodingdefault}{\sfdefault}{bx}{n}

\DeclareMathOperator*{\argmax}{arg\,max}

\usepackage{hyperref}
\usepackage{url}
\usepackage{graphicx}
\usepackage{booktabs}

\title{Confidence Calibration in Vision-Language-Action Models}

\author{\name Thomas Zollo \email tpz2105@columbia.edu \\
      \addr Department of Computer Science\\
      Columbia University
      \AND
      \name Richard Zemel \email zemel@cs.columbia.edu \\
      \addr Department of Computer Science\\
      Columbia University}

\def\month{10}  
\def\year{2026} 
\def\openreview{\url{https://openreview.net/forum?id=OmpTwegbHR}} 

\begin{document}

\maketitle

\begin{abstract}

Trustworthy robot behavior requires not only high levels of task success but also that the robot can reliably quantify how likely it is to succeed.
To this end, we present a first-of-its-kind study of confidence calibration in vision-language-action (VLA) foundation models, which map visual observations and natural language instructions to low-level robot motor commands.
We establish a confidence estimation baseline for VLAs, examine how task success relates to calibration error and how calibration evolves over time, and introduce two lightweight techniques to remedy the miscalibration we observe: prompt ensembles and action-wise Platt scaling.
Our aim in this study is to begin to develop the tools and conceptual understanding necessary to render VLAs trustworthy via reliable uncertainty quantification.

\end{abstract}

\section{Introduction}\label{sec:introduction}

\begin{figure*}[!t]
\vspace{-5pt}
\centering
\includegraphics[width=0.6\textwidth]{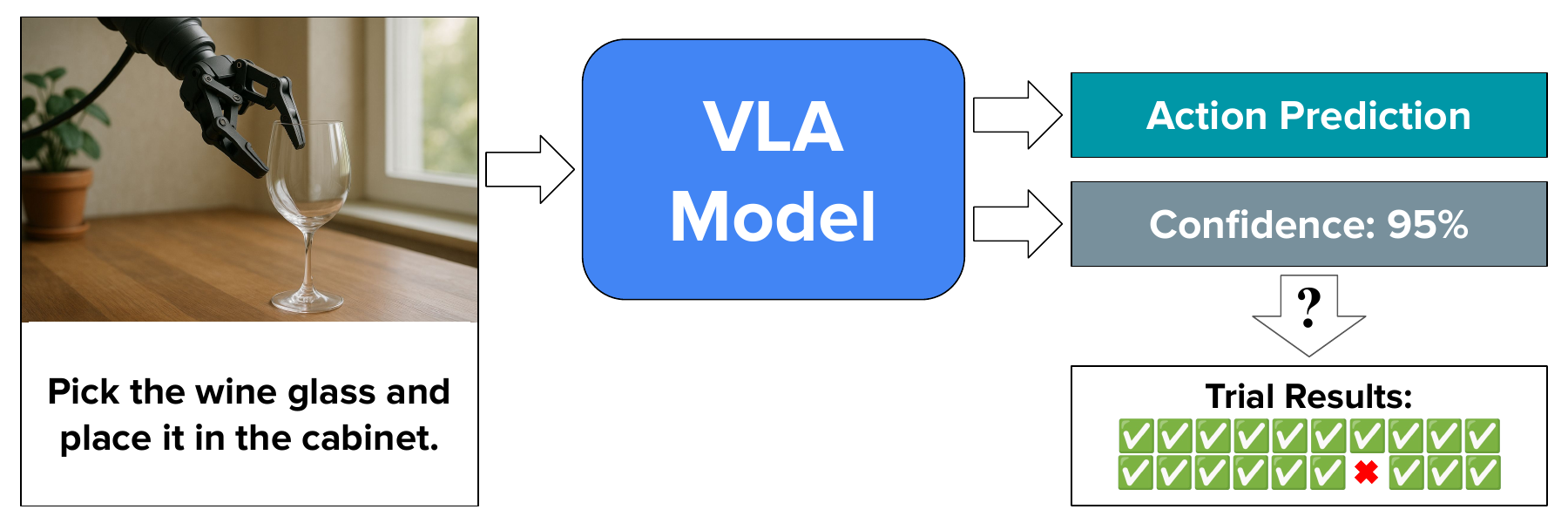}
\caption{
To be trustworthy, a robotic system must be able to reliably express its confidence in its ability to successfully complete a task trial, especially in high-stakes and open-world domains.  
A well-calibrated robot policy produces confidence estimates that align with its probability of trial success, ideally updating its estimate as the task progresses.  
For example, the robot should succeed on 95\% of instances for which it expresses 95\% confidence.
}
\label{fig:main}
\vspace{-5pt}
\end{figure*}

\emph{Confidence calibration}, or the degree to which a model’s predicted probabilities reflect the true likelihood of its predictions being correct, is a cornerstone of reliable machine learning systems \citep{guo2017calibration}.  
When a model is well-calibrated, downstream decision-makers (e.g., humans, planning algorithms, or safety monitors) can trust that a confidence estimate of 95\% implies that the predicted outcome will occur roughly 95\% of the time (see Figure~\ref{fig:main}). 
Mismatches between confidence and actual outcomes can have severe negative consequences in high-stakes settings such as medical diagnosis or autonomous driving, motivating a rich literature on improving calibration in deep learning models \citep{minderer2021revisiting, gal2016dropout, tian2023just, kumar2020verifieduncertaintycalibration, kadavath2022language}.

Recently, the field of robotics has embraced a new class of \emph{vision-language-action} (VLA) foundation models \citep{brohan2023rt2visionlanguageactionmodelstransfer, kim2024openvlaopensourcevisionlanguageactionmodel, black2024pi0visionlanguageactionflowmodel, black2025pi05visionlanguageactionmodelopenworld}. 
Leveraging large-scale, multistage pretraining, these models translate visual observations and natural language instructions into low-level 
robot control
commands, unifying multimodal perception with motor control.
Current VLA systems already demonstrate previously unattainable generalization across environments, tasks, and robot embodiments.
Because these models encode broad semantic and visuomotor priors from large-scale pretraining, they can be efficiently fine-tuned to new robots and downstream tasks, yielding large gains over training from scratch.

Since VLAs are designed for closed-loop interaction with the world, knowing when and how strongly to trust their actions is critical.  
Consider a robot performing a task in a safety-critical environment, or attempting to manipulate a valuable, fragile object.  
In these scenarios, if the policy can accurately express uncertainty, then costly or dangerous accidents can be avoided, such as by refining the instruction or deferring the task to a human.
Despite the importance of calibrated confidence, basic questions remain largely unaddressed, e.g., whether VLAs are calibrated, how calibration evolves over the task horizon, and how it can be improved.

\noindent\textbf{Contributions.}
To bridge this gap, we present a first-of-its-kind study of calibration in VLAs, identifying key open questions and introducing practical remedies for miscalibration.
Our contributions include:
\textbf{(1)} We evaluate the relationship between task success and calibration error across multiple benchmarks and VLA variants, finding that the model architecture and training objective may play a role in determining this relationship.
\textbf{(2)} We establish and validate a simple, model-native confidence-estimation baseline for token-based VLAs by aggregating the probabilities assigned to their selected tokens.
\textbf{(3)} We analyze calibration over task time, identifying natural points for risk-aware intervention.
\textbf{(4)} We propose a lightweight method for ensembling confidence scores across semantically equivalent instructions, consistently cutting calibration error across models and tasks.
\textbf{(5)} We discover systematic over-/underconfidence in different action dimensions and propose a method to recalibrate each action dimension independently to produce more reliable confidence estimates.
Our aim is to develop the tools and conceptual understanding needed to support more trustworthy VLA deployment through reliable uncertainty quantification. Because our findings are focused on fine-tuned, token-based VLAs evaluated in simulation, substantial further work is needed to establish their applicability to broader architectures and real-world settings.

\section{Related Work}\label{sec:related_work}

\paragraph{Confidence Calibration}

A model is considered well-calibrated when the confidence (i.e., probability) it assigns to an outcome matches the long-run frequency of that outcome.
Deviation from this condition is often quantified via
expected calibration error (ECE) \citep{guo2017calibration} while other metrics such as maximum calibration error \citep{guo2017calibration}, Brier score \citep{brier1950verification}, and negative log-likelihood (NLL) are used to capture related and complementary notions of miscalibration.
Despite their impressive accuracy and trends toward better calibration, modern neural networks still display some persistent miscalibration \citep{minderer2021revisiting}, especially when the task distribution does not match the training distribution \citep{ovadia2019trust}.  
Consequently, a rich toolbox of post hoc fixes has emerged: Platt scaling \citep{Platt1999}, temperature scaling \citep{guo2017calibration}, histogram binning \citep{zadrozny2001obtaining}, and other recalibration methods \citep{kumar2020verifieduncertaintycalibration, naeini2015obtaining, zollo2024improvingpredictorreliabilityselective} can be applied to an already trained model to reduce calibration error without altering its decision rule.
Other popular methods to improve calibration of neural networks include dropout \citep{gal2016dropout} and ensembling \citep{lakshminarayanan2017simple}.

\paragraph{Uncertainty Quantification in Robotics}
Uncertainty quantification has long been a central theme in robotics, particularly through the lens of probabilistic robotics \citep{thrun2005probabilistic, deisenroth2011pilco}. 
Recent work has extended this to modern learning-based robot systems, where uncertainty can guide cautious execution, human assistance, exploration, or online adaptation \citep{chua2018deep, pathak2019selfsupervisedexplorationdisagreement, wang2023coplanner, ataei2024dadeewellcalibrateduncertaintyquantification}. 
One direction uses conformal prediction to provide distribution-free guarantees in embodied decision-making. 
For example, \citet{ren2023robotsaskhelpuncertainty} introduce KnowNo, a conformal framework for aligning the uncertainty of LLM-based robot planners to ask for help under ambiguity while maintaining statistical guarantees on task completion. 
Similarly, conformal methods have been used for clarification-seeking in human-robot interaction \citep{lidard2024risk} and online adaptation in imitation learning \citep{zhao2025conformalizedinteractiveimitationlearning}.

A complementary line of work focuses on detecting or managing failures during policy execution. 
\citet{xu2025detectfailuresfailuredata} propose a runtime method for imitation-learning policies that aims to identify failures using only successful training data, while \citet{gu2025safe} introduce SAFE, a failure detector meant to generalize across tasks by leveraging internal VLA representations. 
In other recent work, \citet{valle2025evaluatinguncertaintyqualityvisual} propose several uncertainty and execution-quality metrics for VLA manipulation tasks and evaluate how these metrics correlate with human labels for execution quality. 
Finally, \citet{yuan2026actasklearnuncertaintyaware} study uncertainty-aware policy steering, where a verifier decides whether to execute an action, ask a clarifying language question, or request action-level intervention depending on semantic and low-level action uncertainty.

\section{Calibration in Vision-Language-Action Models}\label{sec:methods}

\begin{figure*}[t]
\vspace{-5pt}
\centering
\includegraphics[width=\textwidth]{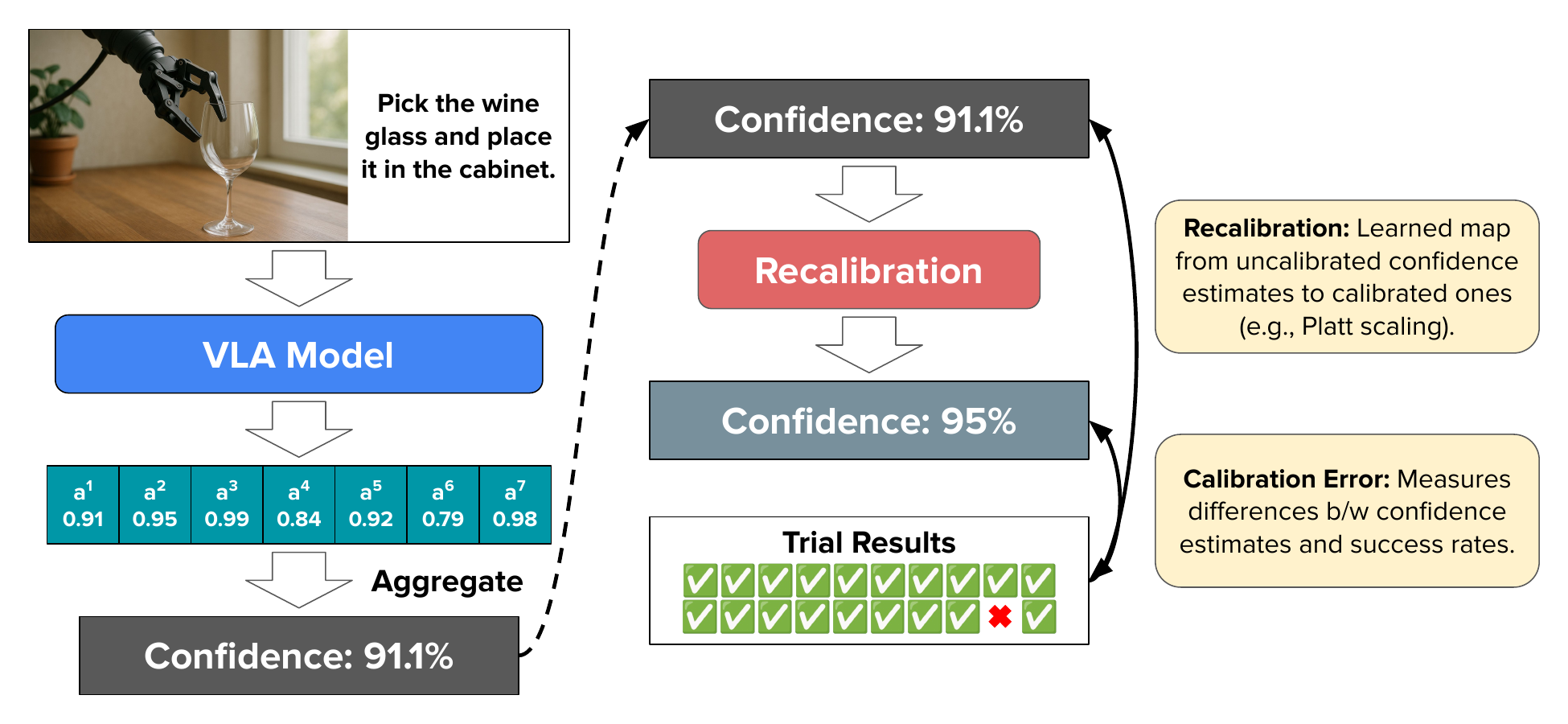}
\caption{
Overview of confidence estimation and calibration for token-based VLAs.
Given an input image and text instruction, a token-based VLA generates distributions over discrete action representations, which may correspond to, e.g., action dimensions, latent action codes, or action chunks depending on the model architecture.
These distributions can be converted into scalar confidence estimates by aggregating the probability assigned to the selected tokens.
Given an uncalibrated confidence estimate, recalibration methods such as Platt scaling use a small calibration dataset to learn a map from uncalibrated confidence estimates to calibrated ones.
Calibration error can then be measured by comparing confidence estimates to actual task success rates.
}
\label{fig:main_2}
\vspace{-5pt}
\end{figure*}

In this section, we formalize the problem of confidence calibration for vision-language-action models. 
We define calibration, describe a typical example of how contemporary VLA architectures generate actions and how to extract a corresponding confidence estimate, and introduce the standard calibration metrics used in our empirical study (see Figure \ref{fig:main_2} for a visual summary of this material). 
Finally, we present two lightweight remedies for the miscalibration we observe in practice: \emph{prompt ensembles} and \emph{action-wise Platt scaling}.  

\subsection{Calibration}\label{subsec:calibration}

Let \(C\!\in[0,1]\) denote the confidence reported by a robot policy and \(Y\!\in\{0,1\}\) the binary indicator of
task success (we will use uppercase for random variables and lowercase for their realizations).  A perfectly calibrated predictor \citep{guo2017calibration, minderer2021revisiting, fisch2022selective} satisfies 
\begin{equation}
    \mathbb{P}(Y=1 \mid C=c)\;=\;c,\qquad\forall\,c\in[0,1].
    \label{eq:perfect}
\end{equation}
If this condition is met, then for the subset of trials on which the robot reports 80\% confidence, we should observe successful completion 80\% of the time.

\subsection{Vision-Language-Action Models}\label{subsec:vla}

\emph{Vision-Language-Action} models take visual data and natural language instructions as input and output robot actions
\citep{brohan2023rt2visionlanguageactionmodelstransfer, kim2024openvlaopensourcevisionlanguageactionmodel, black2024pi0visionlanguageactionflowmodel}. 
They are typically initialized from a visually-conditioned language model (VLM) that, in turn, is initialized from a pretrained language model.
This allows VLAs to leverage rich multimodal priors in joining perception and action generation into a single end-to-end pipeline.
While some VLAs retain the token-based output paradigm inherited from these base models \citep{brohan2023rt2visionlanguageactionmodelstransfer, kim2024openvlaopensourcevisionlanguageactionmodel, lee2025molmoactactionreasoningmodels}, others have augmented the architecture with, e.g., flow matching action experts for smooth, high-frequency control \citep{black2024pi0visionlanguageactionflowmodel, black2025pi05visionlanguageactionmodelopenworld}. 

At task timestep \(t\), a VLA policy \(\pi_\theta\) has access to the observed history
\(o_t = (v_{\le t}, l_{\le t}, a_{<t})\), where 
\(v\) is a visual observation, \(l\) is the (possibly fixed) natural language instruction, and \(a\) is an action.
The policy induces a distribution over the next action $\pi_\theta(a \mid o_t).$
When performing a task, the policy executes the most likely action
$a_t^\star = \argmax_a \pi_\theta(a \mid o_t)$.
Given the need for calibration, we must extract a scalar confidence score \(c_t\) from the policy.
We will interpret \(c_t\) as the policy’s estimate of the probability that the task will ultimately succeed given the observation history and the chosen action, i.e., \(\mathbb{P}(Y=1 \mid o_t, a_t^\star)\).

Next, we propose a baseline method for extracting \(c_t\) for a broad class of token-based VLAs.
We focus on token-based VLAs because they allow for a familiar probabilistic interpretation; we leave it to future work to establish baselines for other families. 
However, our subsequent analysis treats \(c_t\) as a black-box number, so the measures and methods introduced also apply naturally to flow-based, diffusion-based, or other controllers (e.g., those that output action chunks) as long as they can emit such a scalar.

\subsection{Baseline Confidence Estimation for Token-Based VLAs}\label{subsec:baseline}

Many VLAs represent actions using discrete tokens, in close analogy with the token-based outputs of the multimodal language models upon which they are often built.
For example, OpenVLA \citep{kim2024openvlaopensourcevisionlanguageactionmodel} and RT-2 \citep{brohan2023rt2visionlanguageactionmodelstransfer} represent an action with \(D\) discrete tokens, where each token corresponds to one dimension of the robot's action space.
For each dimension \(d\in\{1,\dots,D\}\), the policy outputs logits \(z^{(d)}_t\) and probabilities
\mbox{$
    p^{(d)}_t = \operatorname{softmax}\!\bigl(z^{(d)}_t\bigr)
$},
\mbox{$
    p^{(d)}_{t,k} = \pi_\theta\!\bigl(A^{(d)}_t = k \mid o_t\bigr)
$}
for action tokens \(k\in\{1,\dots,K\}\) in an action vocabulary of size \(K\).
At time \(t\), the policy selects the highest-probability token \(a^{(d)}_t=\argmax_k p^{(d)}_{t,k}\) for each action dimension \(d\), and decodes these tokens into an action for execution.

A common heuristic in LLMs is to use the probabilities assigned to generated tokens as a confidence signal for task success, often measured by answer accuracy \citep{malinin2021uncertainty,kadavath2022language}. 
Analogously, we derive a heuristic confidence signal for VLAs using the probability assigned to each selected action-related token.
In the dimension-wise case above, this corresponds to using \(\max_k p^{(d)}_{t,k}\) for each action dimension \(d\).
Given these selected-token probabilities, we can aggregate them into a single scalar confidence estimate \(c_t\), for instance by taking the mean $c_t \;=\; \frac1D\sum_{d=1}^{D}\max_k  p^{(d)}_{t,k}.$
Other possible aggregation rules over these per-token confidence values include a geometric mean, minimum, maximum, or learned aggregation function.
We compare various aggregation rules empirically in Appendix~\ref{app:agg_comp}; in our experiments, we use the arithmetic mean.

Although action-token probabilities are trained to reflect confidence in the model’s immediate action prediction rather than directly in the final episode outcome, the two can be linked in a closed-loop policy. 
Episodes are more likely to succeed when the policy encounters familiar states and assigns high probability to the actions it selects, whereas low selected-action probability can indicate ambiguity, distribution shift, or uncertainty in perception and control that increases the likelihood of downstream failure. 
We therefore treat aggregated selected-token probability as a model-native proxy for episode-level success and empirically evaluate whether this proxy is calibrated, rather than assuming that the correspondence holds by construction.

Not all token-based VLAs expose action probabilities in this exact dimension-wise form.
Some models predict multiple discrete tokens per action dimension \citep{lee2025molmoactactionreasoningmodels}, while others predict latent action tokens or tokenized action chunks that are decoded jointly into continuous actions \citep{bu2025univlalearningacttaskcentric,hung2025norasmallopensourcedgeneralist}.
In these cases, the exact form depends on the decoder, but the same principle applies: we use the probabilities assigned to the selected action tokens as the primitive confidence signal.

\subsection{Measuring Calibration}\label{subsec:metrics}

Calibration metrics translate deviations from the condition in \eqref{eq:perfect} into quantitative measures of \emph{miscalibration}.  
Given the difficulty of measuring such a condition 
(i.e., estimating conditional success probabilities from finite data),
we consider a range of metrics.
For what follows, let \(\{(c_i,y_i)\}_{i=1}^N\) denote the reported confidence and binary outcome for each of \(N\) robot trials (episodes). 
Each trial \(i\) consists of timesteps \(t=1,\ldots,T_i\) with per-timestep confidences \(c_{i,t}\); to obtain a single trial-level value \(c_i\) we apply an aggregation function \mbox{\(h\): \(c_i = h(\{c_{i,t}\}_{t=1}^{T_i})\).} 
Possible choices include the confidence before the first action, the mean across timesteps, or the min/max over the trajectory. 
The following measures are agnostic to choice of $h$; in our experiments we primarily use the pre-action confidence \(c_i = c_{i,1}\), reflecting the high-stakes open-world robotics setting in which early risk assessment is particularly valuable.

One popular measure of miscalibration is \textbf{expected calibration error (ECE)} \citep{guo2017calibration}:
\begin{equation}
    \text{ECE}_q
    \;=\;
    \bigl(
      \mathbb{E}_{C}\!\bigl[\lvert\mathbb{P}(Y=1\mid C)-C\rvert^q\bigr]
    \bigr)^{\!1/q}\!.
    \label{eq:eceq-pop}
\end{equation}
Put simply, ECE measures the expected difference between confidence and accuracy over the robot's task data distribution.
The parameter $q$ is typically set to $q \in \{1,2\}$.
Because we can observe only a finite sample of trial results, the conditional expectation in \eqref{eq:eceq-pop} cannot be directly measured.
Instead, it is typically approximated with a binning-based estimator.
With results from \(N\) robot trials, we approximate the population ECE quantity in \eqref{eq:eceq-pop} by first ordering predictions according to confidence, and splitting them into \(M\) equal-sized bins \(B_1,\dots,B_M\).  
Then, our empirical estimate $\widehat{\text{ECE}}_q$ is given by
\mbox{$
    \widehat{\text{ECE}}_q
    \;=\;
    \Bigl(
    \sum_{m=1}^{M}\frac{|B_m|}{N}
      \bigl|\mathrm{acc}(B_m)-\mathrm{conf}(B_m)\bigr|^{q}
    \Bigr)^{\!1/q},
$}
where 
$\mathrm{acc}(B_m)$ and $\mathrm{conf}(B_m)$ are the average accuracy and confidence in the bin.

\textbf{Brier score} \citep{brier1950verification} is a classic measure of the quality of probabilistic forecasting:
\mbox{$
    \mathrm{BS} = \frac{1}{N}\sum_{i=1}^{N}(c_i-y_i)^2.
$}
Brier score is an example of a \emph{proper scoring rule} \citep{Gneiting01032007}, meaning that it is minimized in expectation when the predicted probabilities match the true conditional success probabilities.
Brier score rewards both reliability (confidence matching success) and sharpness (predictions away from the population rate).
Another metric used to measure calibration \citep{guo2017calibration} is \textbf{negative log-likelihood (NLL)}:
$
    \mathrm{NLL}
    =-\frac{1}{N}\sum_{i=1}^{N}\!\bigl[y_i\log c_i+(1-y_i)\log(1-c_i)\bigr]
$.
Also a proper scoring rule, NLL penalizes very confident failures much more heavily than Brier score: for a failed trial ($y_i=0$), the Brier penalty $c_i^2$ is bounded by 1, while the NLL penalty $-\log(1-c_i)$ diverges as $c_i\to 1$ (with the analogous result holding for a successful trial as $c_i\to 0$).

\subsection{Prompt Ensembles}\label{subsec:prompt}

Similar to the VLMs from which they are derived
\citep{zhou2024analyzingmitigatingobjecthallucination},
meaning-preserving lexical differences in instructions, e.g.,
\emph{``pick up the coffee cup''} vs.\ \emph{``grab the mug''},
might shift visual attention, alter path planning, and thereby change the
predictions and confidence scores emitted by VLA models.
Although such sensitivity can pose a challenge to reliable task execution,
it also creates an opportunity to employ an ensemble-based approach
\citep{lakshminarayanan2017simple} to confidence estimation.
Specifically, following a model-averaging interpretation, we treat the
particular wording of an instruction as a latent nuisance variable and
average over semantically equivalent rephrasings.
Such averaging can reduce variance in the final confidence estimate by
canceling noise induced by word choice.

Implementing this idea as an algorithm, we:
\textbf{(1) Generate rephrasings.}
An auxiliary LLM produces \(r\) semantically equivalent prompts
\(\mathcal L_{alt}=\{l_{alt}^{(1)},\dots,l_{alt}^{(r)}\}\)
(see Table~\ref{tab:reprompting_example} for examples).
\textbf{(2) Estimate confidence with each variant.}
Generating an action with \(l_{alt}^{(i)}\) yields a confidence
\(c_t^{(i)}\).
\textbf{(3) Aggregate.}
The final estimate is the ensemble mean
\(c^{ens}_t=\frac1r\sum_{i=1}^{r}c_t^{(i)}\).

During deployment, the rollout and all executed actions are generated using
the original instruction; the rephrasings are used only to produce auxiliary
confidence estimates, and their predicted actions are discarded.
The averaged score is therefore intended to estimate episode-level success
for the original-instruction policy, rather than confidence in an averaged or
voted control action.
Conceptually, this prompt ensemble technique can play a similar role to full
model ensembles \citep{lakshminarayanan2017simple} or inference-time dropout
\citep{gal2016dropout}.
The approach is agnostic to the VLA architecture and can also be applied to,
e.g., a flow-based VLA controller, provided that it can produce a scalar
confidence estimate \(c_t\).
With efficient batching on parallel hardware, the method can add little
wall-clock latency in practice.

\subsection{Action-Wise Scaling}\label{subsec:platt}

A standard remedy for miscalibration in classification models is to gather a small validation set from the task distribution and perform post hoc recalibration, learning a function that maps uncalibrated confidence estimates to calibrated ones \citep{Platt1999, naeini2015obtaining, zadrozny2001obtaining, guo2017calibration}.
Post hoc recalibrators are typically simple functions of the original score function (e.g., logits) or confidence estimate.
One popular example of a post hoc recalibration method is Platt scaling \citep{Platt1999, kumar2020verifieduncertaintycalibration}.
Given confidence outputs and binary outcomes \(\{(c_i,y_i)\}_{i=1}^N\) on a held-out validation set,
one fits 
a function
\mbox{\(g(c)=\sigma(\alpha c+\beta)\)} that minimizes NLL,
\mbox{$
    \min_{\alpha,\beta}
        -\sum_i \Bigl[ y_i\log g(c_i) + (1-y_i)\log\!\bigl(1-g(c_i)\bigr) \Bigr]
$},
where \(\sigma(x)=1/(1+e^{-x})\).
At inference each \(c_t\) is replaced with \(\tilde c_t=g(c_t)\),
with the goal of aligning confidence with the probability of task success while leaving the model’s actions unchanged.

Unlike classification models that emit one predictive distribution per sample,
many VLAs output one distribution per
action dimension.
Those dimensions can differ dramatically, for example because gripper open/close appears in nearly 100\% of demonstrations, whereas ``rotate wrist'' is rarer.
One global transform therefore may not be able to correct all dimensions simultaneously.

\noindent\textbf{Action-Wise Platt Scaling.}
We propose to address this heterogeneity by fitting one transform
per dimension \(d\):
\begin{equation}
    \tilde c_t
    \;=\;
    \frac{1}{D}\sum_{d=1}^{D}
        \sigma\!\bigl(\alpha_d\,\max_{k} p^{(d)}_{t,k}
                      +\beta_d\bigr),
    \label{eq:awps}
\end{equation}
with parameters \(\{\alpha_d,\beta_d\}\) learned on the same calibration
set.
Intuitively, \(\alpha_d\) scales dimension \(d\),
flattening or sharpening its distribution, while \(\beta_d\) shifts its
overall optimism.
As in standard Platt scaling, the chosen tokens \(a^{(d)}_t\) remain unchanged; we only modify the reported confidence.
This dimension-wise perspective can be applied to other post hoc recalibrators as well, for instance action-wise temperature scaling.
Beyond the specific methodology, we aim to highlight that VLA calibration will require domain-specific tools rather than a direct transplant of methods designed for standard classifiers.

\section{Experiments}\label{sec:experiments}

Having described the problem of confidence calibration in vision-language-action models, we next conduct an empirical investigation of the following questions:
\textbf{(1)} Does a higher success rate imply better calibration (Section~\ref{sec:exp_tradeoffs})?
\textbf{(2)} How does calibration evolve over the task time horizon (Section~\ref{sec:exp_time})?
\textbf{(3)} Can prompt ensembles consistently improve confidence estimates (Section~\ref{sec:exp_reprompt})?
\textbf{(4)} Are action dimensions differentially calibrated, and what are the implications for recalibration (Section~\ref{sec:exp_dimensions})?
Our aim is to highlight key issues and lay the empirical groundwork for future research on calibrating VLAs.

We perform our experiments using 4 different VLA variants: OpenVLA, MolmoAct \citep{lee2025molmoactactionreasoningmodels}, UniVLA \citep{bu2025univlalearningacttaskcentric}, and NORA \citep{hung2025norasmallopensourcedgeneralist}. 
We choose these models because they represent a variety of VLA design decisions while still all predicting tokens in a manner that allows for a natural probabilistic interpretation.\footnote{Descriptions of confidence estimation with MolmoAct, UniVLA, and NORA are found in Appendix \ref{app:conf_est}.}
All models are fine-tuned on 4 different task suites from the \mbox{LIBERO \citep{liu2023liberobenchmarkingknowledgetransfer}} benchmark: Spatial, Object, Goal, and 10. 
LIBERO is a simulation environment for language-conditioned robot manipulation tasks inspired by human activities (see Table~\ref{tab:dataset_examples} for task examples). 
We deliberately standardize on LIBERO because it is, to our knowledge, the dominant open-source benchmark for modern VLAs and the only one for which a wide range of fine-tuned models are publicly released, enabling comparisons across independently developed VLAs in a common setting;  collecting enough real robot data to meaningfully study expected calibration error would be prohibitively expensive (see Limitations Section~\ref{sec:limit}). We also include results for the 8-bit and 4-bit quantized versions of OpenVLA fine-tuned on Spatial, Object, and Goal, for a total of \textbf{22 model/task suite combinations}.

Each task suite features 10 different tasks with 50 randomized initializations, for a total of 500 examples.  
For a given task suite, task success rate represents the proportion of the 500 trials that result in success (while task error rate represents the proportion of trials that result in failure).
To calculate ECE, we use 12 equal-mass bins and the Python package released with \citet{kumar2020verifieduncertaintycalibration}. 

In Sections \ref{sec:exp_tradeoffs}, \ref{sec:exp_reprompt}, and \ref{sec:exp_dimensions}, we focus on confidence estimates produced at the first timestep, before any action is taken.
This aligns with the need for safety in open-world robot deployments, where robots should signal uncertainty as early as possible in order to avoid costly or dangerous incidents.  
While alternative heuristics, such as averaging confidence across the trajectory, may also be reasonable, they would often incorporate estimates produced after the robot has already failed or entered an unsafe state.
In Section~\ref{sec:exp_time}, we explore how calibration differs across timesteps over the task horizon.
To aggregate action-token-level confidence values into a single scalar estimate, we use the arithmetic mean.
We compare this aggregation rule with alternatives in Appendix~\ref{app:agg_comp}.
Because action-token probabilities are not trained against episode-level outcomes, their correspondence with task success is not guaranteed. 
We therefore view the following experiments partly as an empirical validation of this proxy: consistent alignment with trial outcomes across models and tasks would indicate that selected-token confidence contains useful information about eventual success, while variation across settings would expose its limitations.

\subsection{Relationship Between Calibration and Task Success}\label{sec:exp_tradeoffs}

Much of the early research on calibration in deep learning models was based on perceptions of how task success (usually image classification accuracy) was related to calibration error.
In particular, influential work argued that modern neural networks are poorly calibrated \citep{guo2017calibration}, suggesting that improved accuracy comes at the expense of calibration.
This led researchers to propose training interventions to augment the cross-entropy loss with other objectives seeking improved calibration, potentially at a cost to overall accuracy \citep{pmlr-v80-kumar18a, mukhoti2020calibratingdeepneuralnetworks}. 
However, subsequent research found that, in fact, more accurate networks are generally better calibrated (and easier to recalibrate), and thus modifications to training procedures may not be needed \citep{minderer2021revisiting}.
Instead, techniques such as post hoc recalibration \citep{guo2017calibration, zadrozny2001obtaining, kumar2020verifieduncertaintycalibration} and ensembling \citep{lakshminarayanan2017simple, fort2024ensembleeverywheremultiscaleaggregation}, applied to models trained for high accuracy, are sufficient to achieve low calibration error (though calibration under domain shift remains a challenge \citep{ovadia2019trust}).

\begin{figure*}[t]
\vspace{-5pt}
\centering
\includegraphics[width=\textwidth]{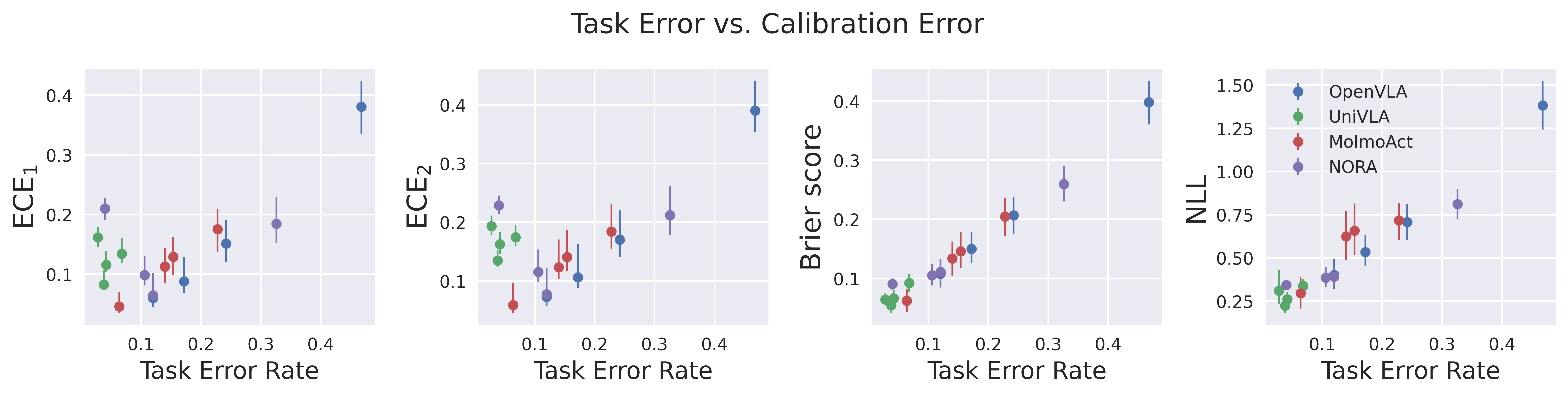}
\caption{
Visualization of task error rates compared against 4 different calibration error measurements for 4 VLA variants (OpenVLA, MolmoAct, UniVLA, and NORA) and 4 LIBERO task suites (Spatial, Object, Goal, 10), as well as OpenVLA 8- and 4-bit versions on Spatial, Object, and Goal.  
Vertical error bars report 95\% percentile-bootstrap intervals for the calibration metrics obtained from
2,000 episode-level resamples with replacement within each model--task-suite
setting.
All models exhibit a roughly monotonic relationship between task error and the discriminative measures (Brier score and NLL).  ECE shows differences between model families, potentially due to architecture and objective differences. 
}
\label{fig:tradeoff_baseline}
\vspace{-5pt}
\end{figure*}

To establish high-level direction for calibration research in VLAs, our first experiment focuses on this important question of how task success relates to pre-execution calibration error (according to $\text{ECE}_1$, $\text{ECE}_2$, Brier score, and NLL) across the 22 model/task suite combinations described above.  
Confidence estimates are produced using the baseline methods described in Section~\ref{sec:methods} and Appendix \ref{app:conf_est}.
Results are visualized in Figure~\ref{fig:tradeoff_baseline}, and also reported in Table \ref{tab:all_results}.
All models exhibit a roughly monotonic relationship between task error and the discriminative metrics (Brier score and NLL), but they differ on ECE: OpenVLA and MolmoAct tend to achieve lower ECE when task error is low, whereas UniVLA and NORA do not show as clean a trend. 
One possible explanation is that in OpenVLA and MolmoAct, the cross-entropy loss (a proper scoring rule) more directly supervises discretized action dimension tokens, so bin-averaged confidence better tracks per-task success.
By contrast, the latent or compressed action representations and auxiliary objectives in UniVLA and NORA may introduce a more significant mismatch between token probabilities and success. 
These patterns point to architectural complexity in modern VLAs as a potential source of unfavorable calibration behavior.
Future work is needed to fully understand which design choices drive this behavior.

To determine how much of the observed relationship between task performance and calibration is explained by marginal confidence--success alignment rather than within-suite calibration structure, we present two additional analyses in Appendix~\ref{app:marginal_effects}.
First, we assess marginal calibration by plotting mean first-step confidence against task-suite success rate.
Second, we compare each calibration metric with the value obtained from a constant predictor that assigns every trial the mean confidence of its corresponding model--task suite pair.
These analyses show that alignment between mean confidence and success rate explains part, but not all, of the observed relationship.
OpenVLA and MolmoAct continue to exhibit lower calibration error in higher-performing settings, whereas UniVLA and NORA show weaker and less consistent trends.

\subsection{Calibration Over Task Time}\label{sec:exp_time}

The previous experiment evaluates confidence before the first action is executed, a conservative choice for safety-critical deployments.
Yet many tasks might allow the robot to collect more information without risk before having to express confidence.
For instance, in the wine glass scenario of Figure~\ref{fig:main}, the gripper can hover above the stem, refine its scene representation, and only then decide whether it is confident enough to proceed.
More context should, in principle, yield better calibration.

To test this intuition, we measure calibration across 500 test trials for 100 different levels of task completion ($\{0,1,\dots,99\}\%$), where task completion is calculated as the current timestep index $t$ divided by the total number of timesteps in the task episode (and multiplied by 100).
Because trajectory lengths vary, we align episodes retrospectively by mapping
each target completion percentage to the corresponding timestep within that
episode and computing calibration across episodes at the aligned relative
position. The normalized coordinate depends on the realized episode length and
is therefore not generally observable online; it is used here as an offline
diagnostic to illustrate how confidence quality changes during execution,
while each confidence aggregation rule itself uses only observations available
up to the selected timestep. In deployment, an analogous progress signal could
be calibrated from offline trajectories using elapsed steps or expected task
duration, or inferred online from perception using a task-phase, subgoal, or
object-state estimator.

Results for each level of task completion are averaged across the 500 trials, to study whether high-level confidence and calibration trends might occur.
In this section, we focus on 6 model/task suite combinations (OpenVLA and its 8-bit version, each applied to Spatial, Object, and Goal suites), to understand how any observations might generalize.
Our goal is to examine whether the quality of confidence estimates changes as the robot progresses in its task.
At each measured timestep, we compute $\text{ECE}_1$ and Brier score using the baseline method for producing confidence estimates.
Because a downstream safety monitor might consider the history of recent estimates (beyond the current one), we evaluate three aggregation rules for reporting a confidence estimate at each timestep:
\textbf{Current} - confidence from the current timestep only;
\textbf{Window (5)} - mean over the baseline confidence estimates from the current timestep and the four immediately preceding ones;
\textbf{Avg. All} - mean over the baseline confidence estimates from all steps seen so far.

\begin{figure*}[t]
\vspace{-5pt}
\centering
\includegraphics[width=0.75\textwidth]{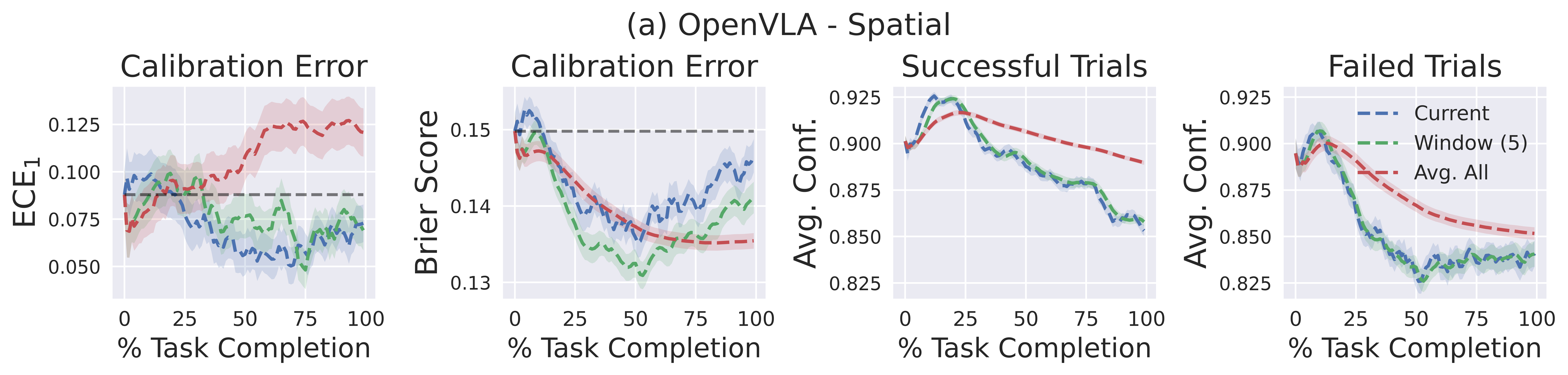}
\includegraphics[width=0.75\textwidth]{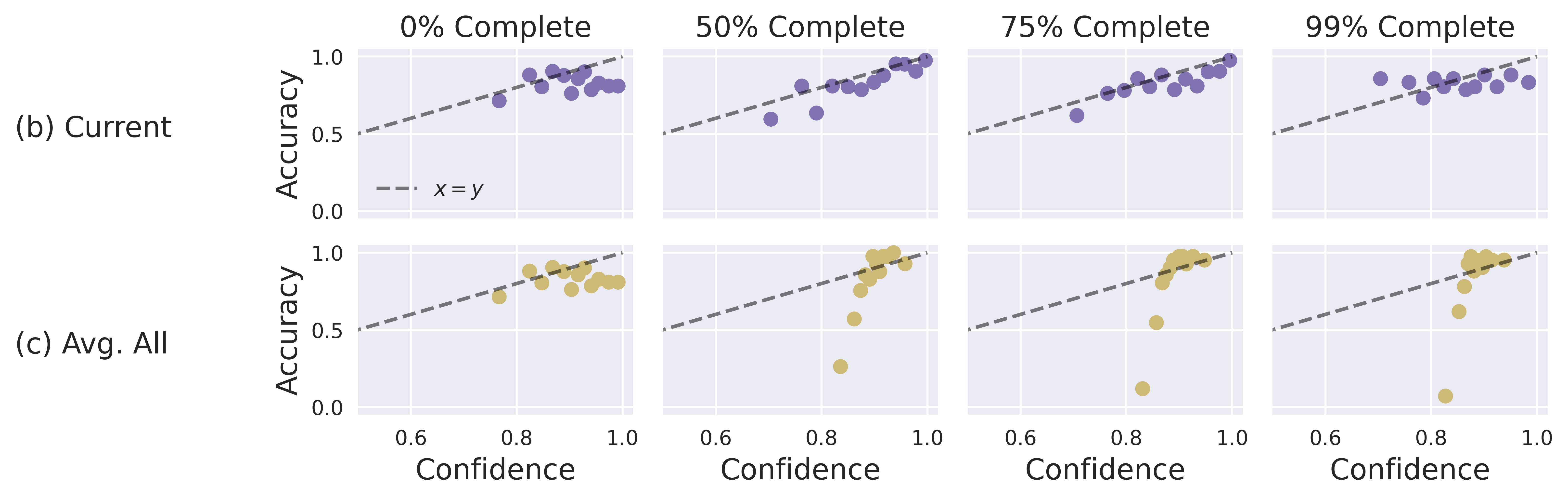}
\caption{Empirical study of calibration error across the task time. 
In the top row (a), the left two plots quantify how calibration evolves with task progress, while the right two plots show the average confidence by task time, grouped by successful and failed trials.
Shaded regions indicate $\pm 1$ standard error: ECE uncertainty is estimated from 500 episode-level bootstrap resamples shared across task-completion points, while Brier and confidence-curve errors are computed analytically across episodes.
The bottom rows provide qualitative reliability diagrams for two representative aggregation rules: current-step confidence (b) and averaging all observed confidences so far (c); we omit the sliding-window variant for visual clarity because its behavior is intermediate between these two. 
Overall, these results illustrate that calibration can improve as the task progresses and more information is gathered, suggesting opportunities for context-aware uncertainty interventions.}
\label{fig:across_time}
\vspace{-5pt}
\end{figure*}

Beyond calibration error, we also visualize how average confidence estimates evolve over the task horizon, separated by successful and failed trials.
Finally, we plot reliability diagrams for the timesteps corresponding to $\{0,50,75,99\}\%$ completion. 
Reliability diagrams \citep{guo2017calibration} offer a visualization of expected calibration error, using a similar binning strategy.
They show confidence against accuracy for each bin, where a well-calibrated system lies on the $x=y$ line. 
Results for the Spatial task suite with the fine-tuned OpenVLA model are shown in Figure~\ref{fig:across_time}.  Additional results for Spatial with the Quant-8 model (Figure \ref{fig:across_time_spatial_quant8}), as well as the Object and Goal suites with the full precision (Figures \ref{fig:across_time_object}, \ref{fig:across_time_goal}) and Quant-8 models (Figures \ref{fig:across_time_object_quant8}, \ref{fig:across_time_goal_quant8}), are presented in Appendix \ref{app:task_time}.

Focusing first on results using the confidence estimate from the current step, a clear pattern emerges.  
As shown in Figure~\ref{fig:across_time} and Figures \ref{fig:across_time_spatial_quant8}-\ref{fig:across_time_goal_quant8}, across all 6 settings and both metrics calibration improves sharply from 0\% to approximately 50\% completion, then plateaus or deteriorates back towards the original level (left-most and left-middle plots in section (a) of the results figures).
Beyond lower scores on calibration error, our additional plots characterize the improved probabilistic nature of the confidence estimates towards the middle of the task horizon.
In the right-middle and right-most plots in the top section (a) (titled ``Successful Trials'' and ``Failed Trials''), the gap between the average confidence on successful vs.~failed trials tends to be greatest around 50\% task completion.
Additionally, considering the top row of reliability diagrams (section (b)), we can see that the confidence estimates using the current step become far more reliable around this point: error for many bins is large at the beginning, then the difference between confidence and accuracy for most bins becomes smaller until the task is roughly 75\% complete.  
Near task completion, the gap between confidence on successful vs.~failed trials closes again, and reliability suffers as a result.
This late-stage increase in calibration error also highlights a limitation of selected-action probability as a proxy for episode-success confidence. 
If an irreversible failure has already occurred, a direct estimate of eventual task
success should approach zero. 
However, the policy may still assign high probability to its immediate next action, including recovery or terminal actions, even when successful task completion is no longer possible.
Consequently, local action certainty can remain high on failed trajectories, causing it to become increasingly misaligned with the episode-level success label near the end of execution. 
Taken together, these results suggest that the usefulness of selected-action confidence is itself progress-dependent: additional perceptual context can make the signal more informative during the middle of a trajectory, while late-stage failures can expose its limitations as a proxy for eventual success.
This motivates the development of progress-aware monitors that condition confidence-based intervention on task progress or phase. 
We leave the design and validation of such monitoring policies to future work.


Next, we observe the effects of other simple strategies for aggregating confidence estimates across timesteps.
As described above, alongside the current step confidence, we also consider a sliding window average over 5 single-step estimates and the average of all observed single-step estimates.
Across all model/task suite combinations, the sliding window method is particularly effective at improving the Brier score around the middle of the task horizon, although ECE is not always improved.
With respect to averaging all single-step estimates so far, we observe that Brier score improves throughout the task, but ECE generally gets worse.
The underlying behavior driving these changes can be observed in the bottom row of reliability diagrams (marked (c)): the ``Avg.~All'' method is successful at assigning relatively lower confidence to failed examples, but also reduces variance in the estimates such that they are highly overconfident in these bins (i.e., for the leftmost bin at 99\% completion, confidence is still high, but accuracy is near zero).
Thus, while this approach fares poorly according to ECE, it is promising in the sense that it enables more effective discrimination between successful and failed trials.

Finally, to probe the generality of these results, we repeat these experiments using the UniVLA model, presented in Appendix Figures \ref{fig:across_time_univla_spatial}, \ref{fig:across_time_univla_object}, and \ref{fig:across_time_univla_goal}.  
We once again find that calibration improves after making some task progress before deteriorating again, suggesting opportunities for context- and risk-aware applications of confidence quantification.


\paragraph{Qualitative Examples}
To illustrate how context-aware confidence monitoring could work in practice, we apply it to a representative pick and place task from the Goal suite.
The task is to ``put the wine bottle on the rack'', a case where the robot should be relatively conservative to avoid breaking glass.  
We consider a naive approach to context-aware monitoring, proposing to halt task performance when both:
\textbf{(1)} the confidence level falls below a threshold set to the 10\% quantile of confidence estimates for that point in the task horizon across all task trials (based on percent completion);
\textbf{(2)} the robot is within a few inches of contacting an object, or already has contacted an object.
Since building a system to detect proximity to objects is beyond the scope of this work, we focus on a qualitative demonstration of this idea.
For each example under examination, we plot current confidence for $\{0, 20, 40, 60, 80, 98\}\%$ completion, as well as the corresponding 10\% quantile risk threshold and an image of the robot environment at that time.

\begin{figure*}[t]
\vspace{-5pt}
\centering
\includegraphics[width=0.925\textwidth]{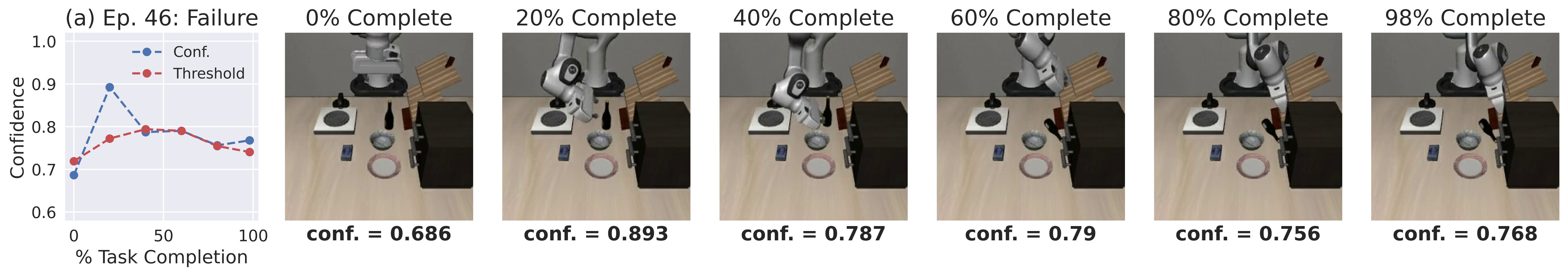}
\includegraphics[width=0.925\textwidth]{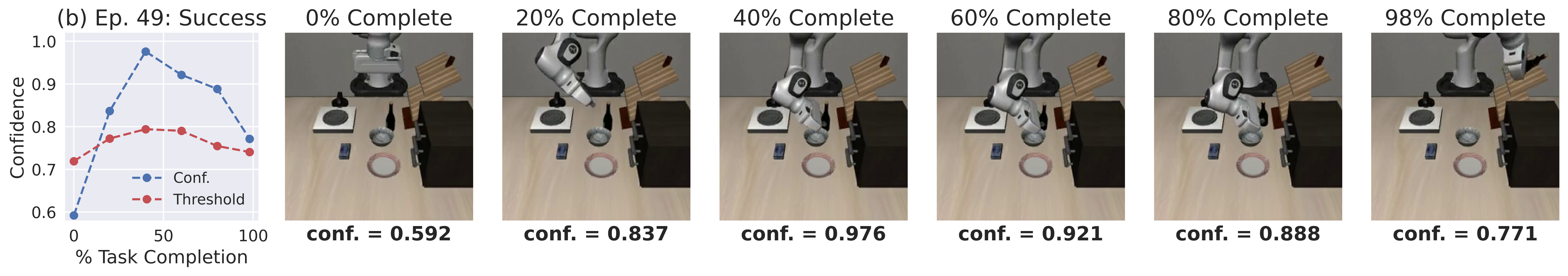}
\caption{
Qualitative examples of a context-aware confidence monitoring strategy applied to a task from the Goal suite.  Here, the task is to ``put the wine bottle on the rack''.  The red dashed line represents the 10\% quantile of the confidence estimates output by the model across the task time horizon, offering a potential threshold below which the robot may abort execution. 
}
\label{fig:qualitative_across_time}
\vspace{-5pt}
\end{figure*}

Some particularly illustrative examples are shown in Figure~\ref{fig:qualitative_across_time}.
In both Episode 46 (marked (a)) and Episode 49 (marked (b)), confidence begins below the threshold. 
However, given the knowledge that calibration improves throughout the task horizon, we may prefer to allow the robot to proceed with the task as long as it is not too near any objects.
Comparing these examples shows the potential of such an approach.  
In Episode 46, 
although confidence increases initially, it drops sharply between 20\% and 40\% task completion
and remains low throughout the rest of the task, as the wine bottle is ultimately knocked down and the model is unable to recover.
Given our proposed strategy, the task might have been halted before the wine bottle was contacted (at 40\% completion), and could, for example, have been deferred to a human.
On the other hand, in Episode 49 confidence estimates remain above the threshold for the rest of the trial, and the robot is successful.
This episode also highlights the usefulness of an adaptive confidence threshold, given that the confidence estimate at 98\% task completion would have fallen below the threshold at other timesteps.
More such qualitative examples are provided in Appendix Figure \ref{fig:qualitative_across_time_appendix}.
These include both cases where the strategy succeeds and others where it fails (e.g., in Episode 1 confidence falls below the threshold while grasping in an ultimately successful trial).

This qualitative experiment is intended as a proof of concept rather than a
validated deployment policy. 
Its purpose is to make the temporal calibration
result operational by illustrating how a progress-conditioned threshold could
alter concrete intervention decisions, while examples in which the heuristic
succeeds and fails expose both its potential and its limitations. 
A practical
online implementation could, e.g., replace retrospective normalized completion with an observable progress estimate calibrated from offline trajectories or inferred from the current perceptual state.

\subsection{Ensembling Confidence Across Prompts}\label{sec:exp_reprompt}

We next study the empirical effectiveness of the prompt ensemble approach described in Section~\ref{sec:methods}.  
First, for the natural language instruction associated with each task, we create 20 rephrasings using GPT-4o-mini (see Appendix Table \ref{tab:reprompt_prompts} for prompts used). 
During testing, we produce a confidence estimate conditioned on each ``reprompt'', and average over these confidence scores to obtain the final model confidence. 
We evaluate our method using OpenVLA across 3 task suites, as well as UniVLA, NORA, and MolmoAct on the Spatial suite.
We measure $\text{ECE}_1$, $\text{ECE}_2$, Brier score, and NLL, comparing to the baseline method for producing confidence estimates using the original instruction.
The original instruction determines the executed rollout and success label, while rephrasings are used only to produce auxiliary confidence estimates; we study action agreement across prompt variants in Appendix~\ref{app:reprompt_action_agreement}.

\begin{figure}[!t]
\vspace{-5pt}
\centering
\includegraphics[width=0.8\columnwidth]{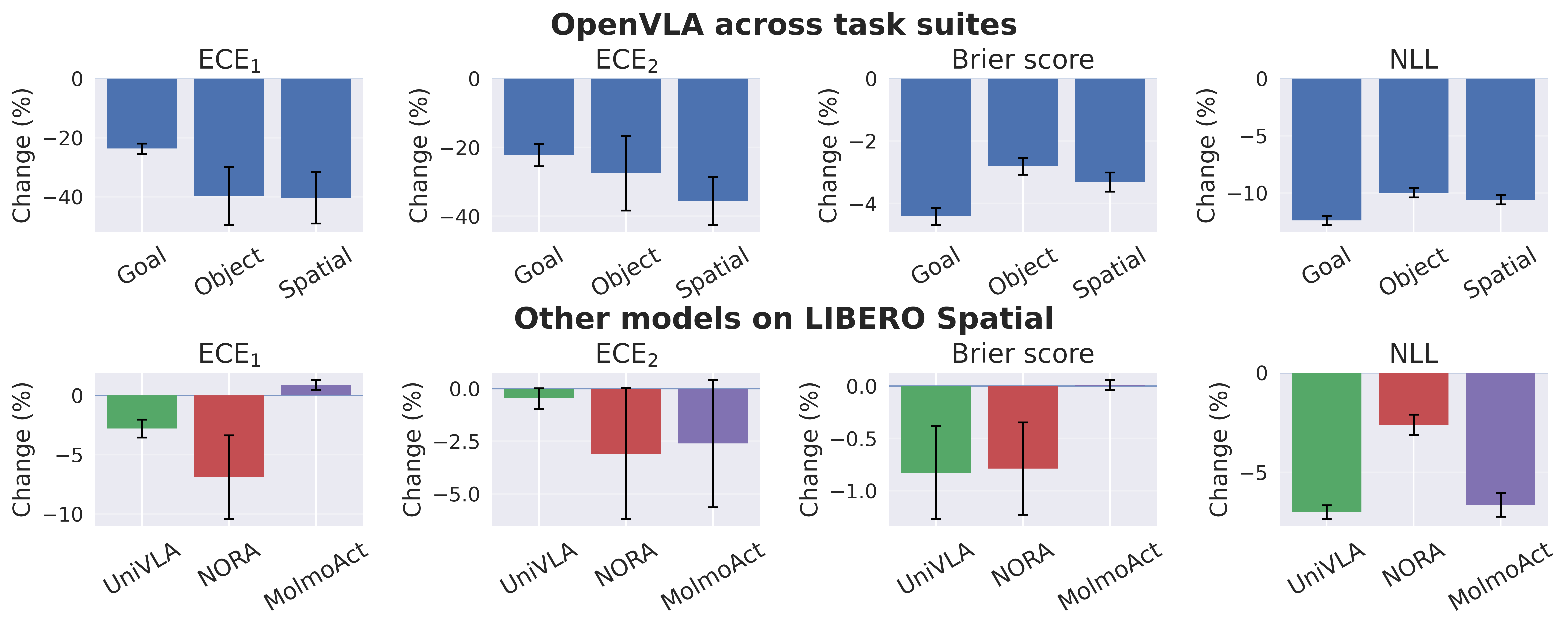}
\caption{ Each bar shows the percent change in a calibration metric after applying Reprompt, relative to the baseline confidence estimate from the original instruction; negative values indicate improved calibration. Results are grouped by OpenVLA task suite and by VLA model on the Spatial suite.
Error bars denote $\pm 1$ prompt-bootstrap standard error, obtained by
resampling the 20 instruction rephrasings.
$\text{ECE}_2$ and NLL always decrease, while $\text{ECE}_1$ and Brier score decrease in 5 of 6 cases.
}
\label{fig:reprompt_reductions}
\vspace{-5pt}
\end{figure}

Detailed results (including for quantized OpenVLA) are recorded in Appendix Table~\ref{tab:appendix_calibration}, while Figure~\ref{fig:reprompt_reductions} summarizes the impact of Reprompt via percent changes across metrics.
To quantify sensitivity to the particular composition of the prompt ensemble,
Figure~\ref{fig:reprompt_reductions} also reports prompt-bootstrap uncertainty, where we resample the 20 rephrasings with replacement and recompute the ensemble confidence and calibration metric.
Across all model/task suite settings shown, Reprompt always decreases $\text{ECE}_2$ and NLL; $\text{ECE}_1$ and Brier score decrease in 5 of 6 cases, with a small increase for MolmoAct on Spatial.
The largest gains occur for OpenVLA across task suites, where ECE reductions can reach roughly 40\%. 
Overall, the improvement is larger for ECE than for the proper scoring rules
(Brier score and NLL), suggesting that gains in reliability (i.e., confidence
matching marginal success rates) tend to exceed those in sharpness, which is
expected from an ensemble technique targeted at reducing variance in
confidence estimation.

The prompt-bootstrap analysis indicates that the OpenVLA improvements are
stable with respect to which rephrasings are included in the ensemble. For the
other architectures, the NLL reductions are also comparatively stable, while
the smaller changes in ECE and Brier score are more sensitive to prompt
composition.
To further characterize robustness, Appendix~\ref{sec:ablations} reports
results across independently generated prompt sets and different ensemble
sizes.

\subsection{Calibration Across Action Dimensions}\label{sec:exp_dimensions}

Some token-based VLAs such as OpenVLA or MolmoAct decompose a low-level command
into tokens explicitly corresponding to their 7 different action dimensions.
Our baseline confidence estimate collapses this structure into a single scalar by averaging the top-token probabilities across the dimensions; 
implicitly, this assumes that every dimension is calibrated to the same
degree.
That assumption may not hold in practice: a gripper ‘open’ or ‘close’ token may appear in nearly every demonstration, whereas a $90^{\circ}$ wrist roll could be rare.
Such dataset imbalances and other implementation details might skew calibration across dimensions, and relevant algorithms, e.g., for post hoc recalibration, might benefit from explicitly addressing this.
To study this question, our final experiment consists of two parts.
\textbf{(1) Per-dimension calibration audit.} For each dimension $d$ we treat $\max_k p^{(d)}_{t,k}$ as that dimension’s confidence and compute $\text{ECE}_1$, to examine whether different degrees of freedom are differentially calibrated.
\textbf{(2) Targeted recalibration test.}  We compare classic Platt scaling to the action-wise variant of Section~\ref{sec:methods}, which learns independent scaling parameters for each action dimension before averaging the transformed confidences across dimensions.

\begin{figure*}[t]
\vspace{-5pt}
\centering
\includegraphics[width=0.49\textwidth]{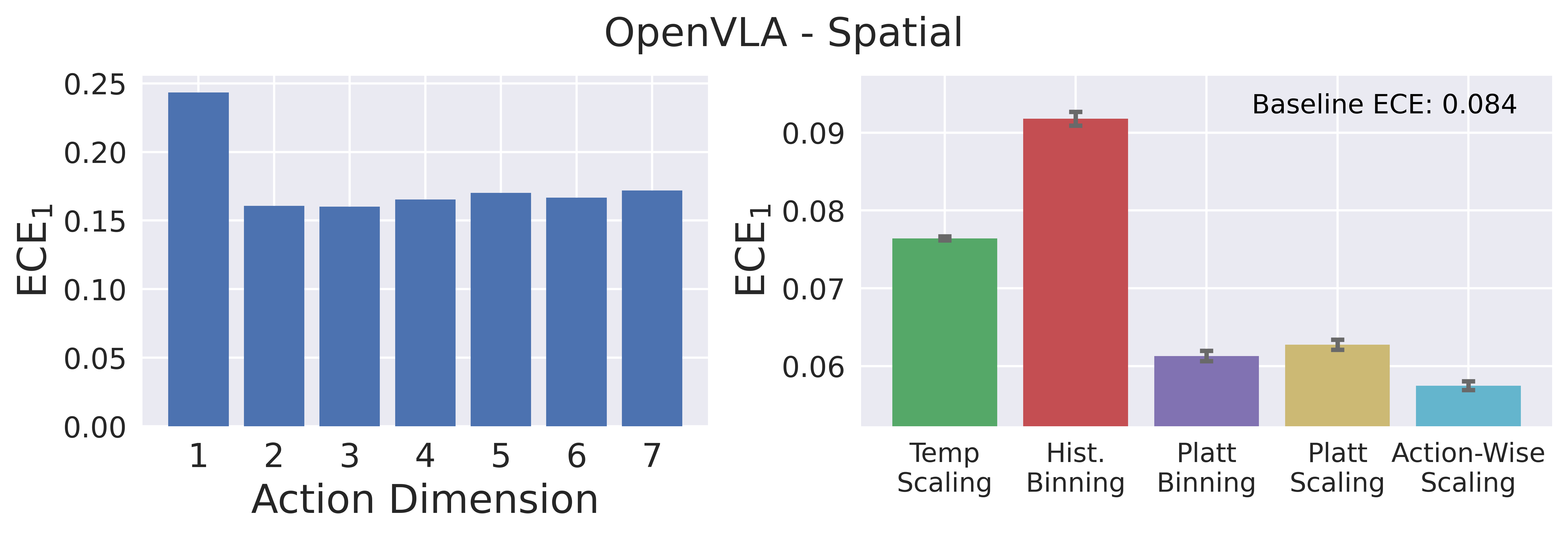}
\includegraphics[width=0.49\textwidth]{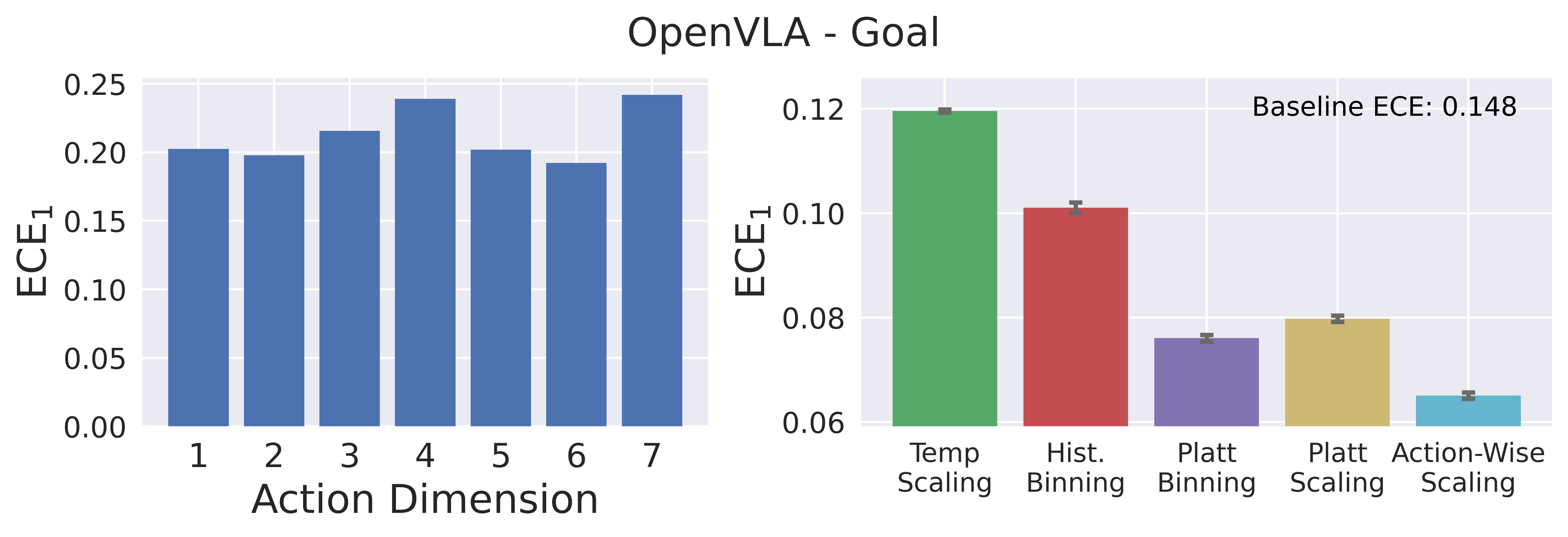}
\includegraphics[width=0.49\textwidth]{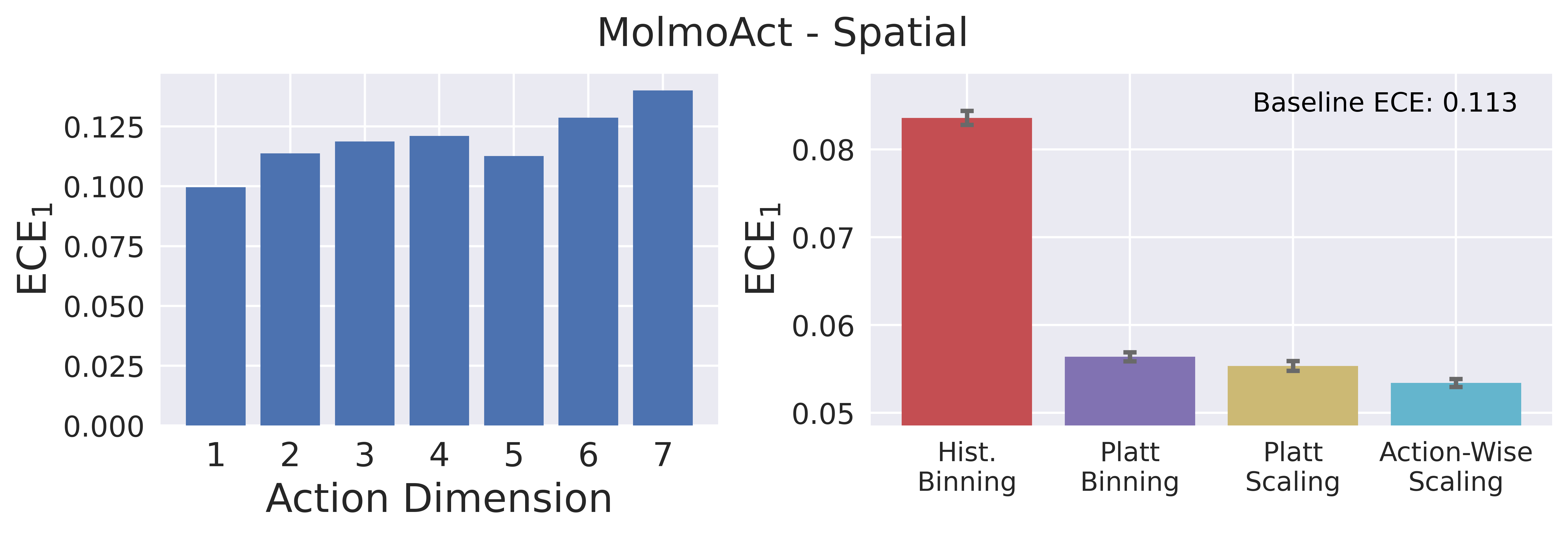}
\includegraphics[width=0.49\textwidth]{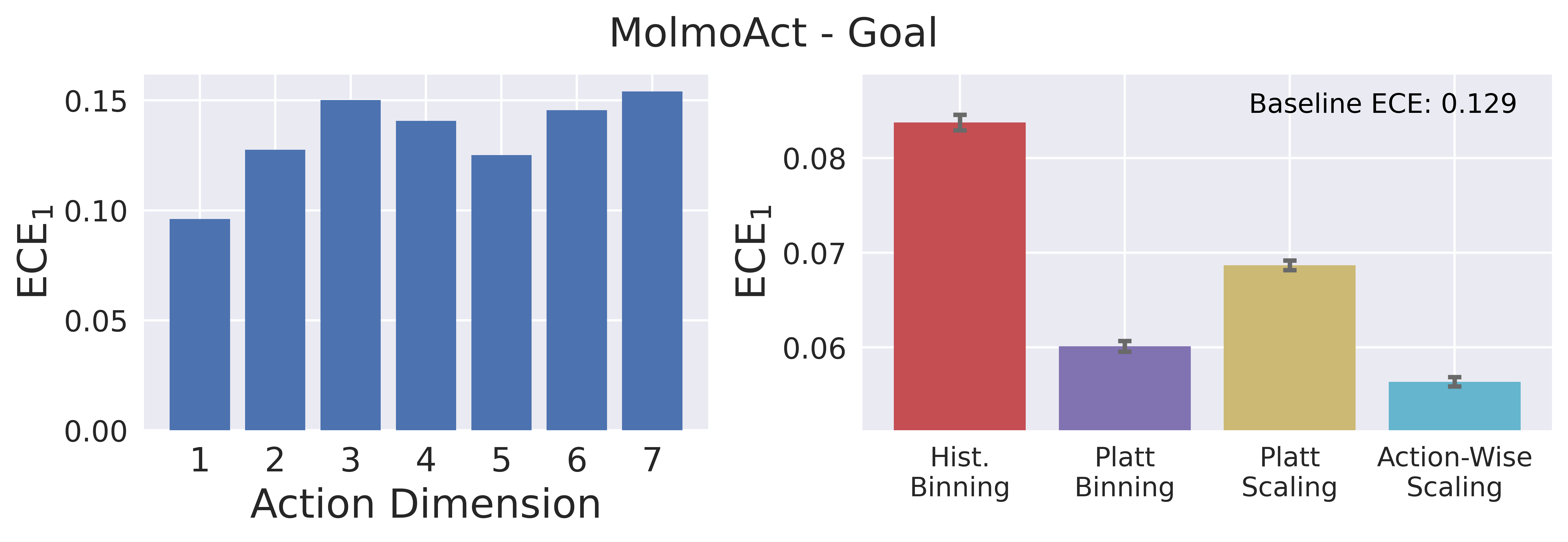}
\caption{
For each model/task setting, the left plot shows ECE for each action dimension using the baseline per-dimension confidence, while the right plot compares recalibration methods (temperature scaling is excluded from MolmoAct results, as the variable-length action encodings preclude a straightforward adaptation). Lower values indicate better calibration in both plots. 
Error bars in the right plots report $\pm 1$ standard error
across the 1,000 random calibration--test splits.
Calibration varies substantially across action dimensions, and action-wise Platt scaling consistently improves over global recalibration baselines.
}
\label{fig:scaling_main}
\vspace{-5pt}
\end{figure*}

We perform our study using MolmoAct and all 3 OpenVLA variants (full, Quant-8, Quant-4) fine-tuned on the Spatial and Goal task suites, as these models output dimension-wise action predictions.
For each setting, we repeat the evaluation over 1,000 random 20\%/80\% calibration/test splits.
Given the reduced size of the test set, we measure calibration with 10 equal-mass bins (instead of 12 as in other experiments).
Beyond traditional Platt scaling, we also include 
a range of baselines: temperature scaling \citep{guo2017calibration}, histogram binning \citep{zadrozny2001obtaining}, and Platt binning \citep{kumar2020verifieduncertaintycalibration}. (Note: temperature scaling is excluded from MolmoAct results, as the variable-length action encodings preclude a straightforward adaptation.)
Figure~\ref{fig:scaling_main} shows results with the baseline estimates for the full precision OpenVLA and MolmoAct models; Appendix Figure~\ref{fig:scaling_appendix} includes the results with the quantized models.

Two key observations arise. First, $\text{ECE}_1$ varies by up to 2 times across dimensions, with no consistent best or worst dimensions across settings.  It follows that a single scalar confidence masks significant differences in dimension-wise confidence estimates.  Second, we can observe that replacing global Platt scaling with per-action-dimension transforms consistently lowers calibration error (and also improves over the additional baselines).  ECE is improved by almost 20\% in some cases, without changing the selected tokens or adding meaningful runtime overhead. Together, these results indicate that VLA calibration research should treat each degree of freedom as its own concern in order to improve the effectiveness of relevant algorithmic tools.
Dimension-aware post hoc methods such as action-wise Platt scaling offer a simple yet effective path toward that goal.
More broadly, our findings underscore that calibrating VLAs introduces challenges not seen in other domains, so meaningful progress will demand substantial domain-specific research rather than wholesale adoption of understanding and techniques from other areas of deep learning.

\section{Discussion}

Our results suggest that VLAs can already provide informative confidence signals, even when confidence is extracted using simple token-probability baselines. In several settings, these estimates achieve relatively low calibration error, suggesting that the probabilistic structure of current VLA policies is already useful for estimating task success. However, calibration varies across architectures, task suites, timesteps, and action dimensions, so low aggregate ECE in one setting should not be interpreted as a general guarantee of calibration. Rather, these findings motivate evaluating VLA confidence as a property of the full model-task-execution pipeline rather than as a fixed property of the policy alone.
The calibration tools we propose provide initial evidence that this structure can be exploited to improve calibration. Prompt ensembling reduces calibration error by averaging over semantically equivalent instructions, suggesting that some variation in confidence arises from lexical sensitivity rather than task-relevant uncertainty. Action-wise Platt scaling improves over global recalibration by accounting for heterogeneity across control dimensions, while the temporal analysis shows that confidence estimates may become more reliable after partial task progress. Together, these results suggest that effective VLA calibration will benefit from methods that account for the linguistic, temporal, and action-structured nature of embodied decision-making.

\section{Limitations}\label{sec:limit}

Our experiments are conducted entirely in simulation using fine-tuned,
predominantly in-distribution VLA policies. Simulation is common in related
robot-learning research
\citep{mees2022matterslanguageconditionedrobotic,
jiang2023vimageneralrobotmanipulation,
li2024masteringrobotmanipulationmultimodal,
li2024visionlanguagefoundationmodelseffective}
and provides the controlled resets and hundreds of rollouts per task suite
needed to estimate calibration reliably. Conducting an equivalent study on
physical hardware would require thousands of robot executions and would be
substantially more time-consuming and expensive. Nevertheless, simulation does
not reproduce important deployment factors such as sensor and actuation noise,
latency, hardware failures, physical damage, or unexpected interactions with
open-world environments. Our results therefore characterize calibration within
the evaluated simulation settings and should not be interpreted as evidence
that these policies are safe or trustworthy in real-world deployment.
Our evaluation is also restricted to the LIBERO benchmark. LIBERO enables
controlled comparisons across multiple independently developed models and
provides sufficient repeated trials for calibration analysis, but it represents
a limited family of simulated language-conditioned manipulation tasks,
environments, and success criteria. 
We additionally focus on token-based VLAs because their discrete predictions
provide a natural likelihood-derived confidence signal. Although the
calibration metrics and general post hoc framework apply to any policy that
emits a scalar confidence estimate, our selected-token baseline and action-wise
recalibration method have not been validated for continuous-action,
diffusion-based, or flow-matching controllers. Furthermore, calibration is only
one component of a trustworthy robotic system: a well-calibrated policy may
still fail frequently, behave unpredictably under distribution shift, or
produce actions that violate physical safety constraints. Extending this work
to real robots, zero-shot and out-of-distribution evaluation, other benchmarks
and environments, broader robot embodiments, and a wider range of VLA
architectures is therefore essential for determining how well these findings
generalize and for developing a more complete account of VLA reliability.

\section{Future Work}\label{sec:future_work}

Several extensions follow naturally from our work.  
Confidence calibration should be benchmarked across other environments, robot embodiments, and VLA architectures, including diffusion-based planners, continuous regression heads, and flow-matching controllers. 
Establishing strong baselines for these architectures will likely require new confidence surrogates and post hoc adjustments. 
Likewise, perturbation techniques besides instruction paraphrasing, e.g., inserting synthetic lighting changes or random distractors, could expose additional failure modes and inspire new multimodal ensemble techniques. 
Running such experiments on real robots will be essential for understanding how various physical factors affect VLA calibration in the real world.
Uncertainty estimation also holds the potential to guide efficient data gathering and fine-tuning, as in active learning \citep{Wang_2017}.
Further, while all of our experiments focused on fine-tuned in-distribution settings, future work should consider these phenomena in the zero-shot and out of distribution settings.
Finally, our temporal analysis hints at a period mid-trajectory where confidence is most trustworthy; integrating that signal into selective execution, planning, or human-in-the-loop systems is an open challenge for designing adaptive risk-aware robotic pipelines.

\section*{Acknowledgements}
We thank ONR Grant N00014-23-1-2436 for its generous support.  
This work is supported by the funds provided by the National Science Foundation and by DoD OUSD (R\&E) under Cooperative Agreement PHY-2229929 (The NSF AI Institute for Artificial and Natural Intelligence).

\bibliography{main}
\bibliographystyle{tmlr}

\clearpage

\appendix

\section{Additional Related Work}

\paragraph{Calibration in LLMs}
Given that most state-of-the-art VLAs are built on LLM backbones, it is natural to consider how our work relates to existing studies of calibration in LLMs, an area that has received considerable attention recently.
Typical approaches to generating confidence estimates with LLMs include expressing calibration as a multiple-choice question for the LLM \citep{kadavath2022language}, directly verbalizing confidence with the model's text output \citep{lin2022teaching, tian2023just, band2024linguistic}, or measuring semantic consistency across many sampled outputs for the same query \citep{kuhn2023semantic, duan2024shifting, chen2024inside}.
However, it is difficult to adapt these methods to the VLA setting. 
For example, while LLMs may produce the entire sequence before measuring confidence, in robotics potential failures must be flagged much earlier in the trajectory to ensure safety and avoid costly accidents.
Also, sampling multiple full trajectories in the physical world may be impossible.
Finally, current VLAs lack the flexible and robust text-to-text interface of LLMs, and thus cannot be expected to, e.g., answer natural language questions about their confidence in an action prediction.

\section{Additional Experiment Details}

This section contains additional experiment details.
All VLA models and benchmarks used in our experiments are open source and publicly available.
Task examples from the LIBERO task suites are shown in Table~\ref{tab:dataset_examples}.

Our code is available at \url{https://github.com/thomaspzollo/vla_calibration}.

\subsection{Confidence Estimation with VLA Variants}\label{app:conf_est}

\paragraph{MolmoAct}
At each timestep, MolmoAct predicts a sequence of discrete action tokens for every action dimension, with a softmax distribution over the 256 action bins for each token. For a given action dimension, we take the probability assigned to the executed bin for each of its tokens and average these probabilities to obtain a per-dimension confidence. This yields one scalar confidence value per action dimension. We then average these per-dimension confidences across all action dimensions to obtain a single scalar confidence for the action at that timestep.

\paragraph{UniVLA}
At each timestep, UniVLA predicts a fixed number (4) of discrete latent action tokens, each chosen from a codebook of size 16.
For each latent token, we take the softmax probability assigned to the selected code as its token-level confidence.
Because these latent tokens are jointly decoded into all continuous control dimensions, the confidence scores do not correspond to particular action dimensions.
Finally, we average these per-token confidences to obtain a single scalar confidence for UniVLA at that timestep.

\paragraph{NORA}
At each timestep, NORA predicts an action chunk encoded as a sequence of discrete tokens, each selected via a softmax over the augmented action vocabulary.  For every token, we take the softmax probability of the sampled token as its token-level confidence. Because the NORA tokenizer mixes information from all control dimensions, we do not derive per-dimension confidences. Instead, we compute a single scalar confidence for NORA by averaging these token-level confidences across all tokens in the predicted action chunk for that timestep.

\begin{table*}[t]
\centering
\caption{Task examples from the LIBERO task suites: Spatial, Object, Goal, and 10.}
\begin{tabular}[t]{p{1.2cm} p{2.5cm} p{9cm}}
\toprule
Dataset & Example Image & Example Instruction \\
\midrule
Spatial & \includegraphics[width=\linewidth]{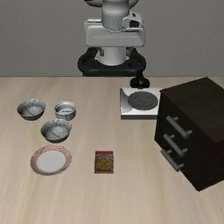} & Pick up the black bowl between the plate and the ramekin and place it on the plate. \\
\midrule
Object & \includegraphics[width=\linewidth]{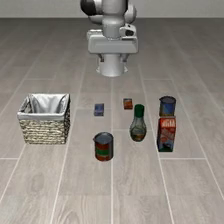} & Pick up the alphabet soup and place it in the basket. \\
\midrule
Goal & \includegraphics[width=\linewidth]{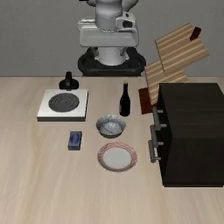} & Open the middle drawer of the cabinet. \\
\midrule
10 & \includegraphics[width=\linewidth]{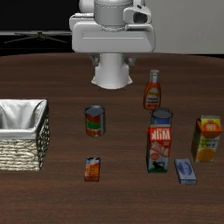} & Put both the alphabet soup and the tomato sauce in the basket.\\
\bottomrule
\end{tabular}
\label{tab:dataset_examples}
\end{table*}

\begin{table*}[t]
\centering
\caption{Example of multiple paraphrases of a single VLA instruction (from our experiments with LIBERO Spatial).  The rephrasings carry the same meaning as the original instruction, enabling a \textit{reprompting} approach in which we ensemble over predictions conditioned on lexically different yet semantically equivalent instructions.}
\begin{tabular}{p{2cm} p{9cm}}
\toprule
\textbf{Instruction} 
& 
Pick up the black bowl between the plate and the ramekin and place it on the plate
\\
\midrule
\textbf{Rephrasings}
& (1) Lift the black bowl located between the plate and the ramekin and set it on the plate. \\
& (2) Grasp the black bowl found between the plate and the ramekin and move it to the plate. \\
& (3) Take the black bowl positioned between the plate and the ramekin and position it on the plate. \\
\bottomrule
\end{tabular}
\label{tab:reprompting_example}
\end{table*}

\subsection{Rephrasing Prompts}

The prompts given to GPT-4o-mini for rephrasing instructions from the LIBERO robot simulation environment are listed in Table ~\ref{tab:reprompt_prompts}.  Prompt 1 is the main prompt used throughout the experiments, and Prompts 2 and 3 are used to ablate the sensitivity to the rephrasings.  Note that the different prompts lead to substantially different rephrasings, as Prompts 2 and 3 mandate the retention of the words ``pick'' and ``place'', while as Table~\ref{tab:reprompting_example} shows, Prompt 1 leads to these words being replaced.

\begin{table*}[ht]
    \centering
    \caption{Prompts for GPT-4o-mini to rephrase LIBERO task instructions.  These prompts are used for Spatial and Object.  There is a small modification for Goal.}
    \begin{tabular}{p{0.2\linewidth}p{0.75\linewidth}}
    \toprule
     & Prompt \\
    \midrule
    Prompt 1 & You are generating alternative phrasings of a robotic task instruction while preserving its exact meaning.
    
        \#\#\# Task Instruction:
        '[TASK DESCRIPTION]'
    
        \#\#\# Instructions:
        - Generate **20** alternative ways to phrase the task instruction.
        - Keep each instruction **concise and unambiguous**.
        - Ensure the instructions remain suitable for a **robot, not a human**.
        - Only make **semantically meaningless** changes (e.g., word order, synonyms, slight rewording).
        - Double-check that the new instructions mean the same exact thing for the robot; do not just substitute synonyms without considering context.
        - Do **not** introduce additional steps, remove essential details, or alter the action.

        \#\#\# Output Format:
        Each rephrased instruction should be wrapped in `[instruction]` and `[/instruction]` tags, like this:
        [instruction] Rephrased instruction 1 [/instruction] 
        [instruction] Rephrased instruction 2 [/instruction]
         \\
    Prompt 2 & You are generating alternative phrasings of a robotic task instruction while preserving its exact meaning.
    
        \#\#\# Task Instruction:
        '[TASK DESCRIPTION]'
    
        \#\#\# Instructions:
        - Generate **20** alternative ways to phrase the task instruction.
        - Keep each instruction **concise and unambiguous**.
        - Ensure the instructions remain suitable for a **robot, not a human**.
        - Only make **semantically meaningless** changes (e.g., word order, synonyms, slight rewording).
        - Double-check that the new instructions mean the same exact thing for the robot; do not just substitute synonyms without considering context.
        - Do **not** introduce additional steps, remove essential details, or alter the action.
        
        - The first word of the instruction should be `PICK', and then it should also include the word `PLACE'.
    
        \#\#\# Output Format:
        Each rephrased instruction should be wrapped in `[instruction]` and `[/instruction]` tags, like this:
        [instruction] Rephrased instruction 1 [/instruction] 
        [instruction] Rephrased instruction 2 [/instruction]
         \\
    Prompt 3 & You are generating alternative phrasings of a robotic task instruction while preserving its exact meaning.
    
        \#\#\# Task Instruction:
        '[TASK DESCRIPTION]'
    
        \#\#\# Instructions:
        - Generate **20** alternative ways to phrase the task instruction.
        - Keep each instruction **concise and unambiguous**.
        - Ensure the instructions remain suitable for a **robot, not a human**.
        - Only make **semantically meaningless** changes (e.g., word order, synonyms, slight rewording).
        - Double-check that the new instructions mean the same exact thing for the robot; do not just substitute synonyms without considering context.
        - Do **not** introduce additional steps, remove essential details, or alter the action.
        - Make the changes as minor as possible, as the robot's language system is not very robust to rephrasing.
        - The first word of the instruction should be `PICK', and then it should also include the word `PLACE'.
    
        \#\#\# Output Format:
        Each rephrased instruction should be wrapped in `[instruction]` and `[/instruction]` tags, like this:
        [instruction] Rephrased instruction 1 [/instruction] 
        [instruction] Rephrased instruction 2 [/instruction]
         \\
    \bottomrule
    \end{tabular}
    \label{tab:reprompt_prompts}
\end{table*}

\clearpage

\section{Additional Experiment Results}

All task success and calibration results for 22 VLA/task suite combinations are shown in Table \ref{tab:all_results}.

\subsection{Comparison of Aggregation Methods}\label{app:agg_comp}

We use the arithmetic mean of selected action-token probabilities as the default confidence score in our experiments. 
In Table~\ref{tab:aggregation_comparison}, we compare this strategy to taking the geometric mean, maximum, or minimum of the same set of token probabilities. 
While the geometric mean is often used as a length-normalized confidence score, we find it to be slightly less effective for VLA calibration on average across our testbed. 
This may reflect the fact that downstream task success is not equivalent to the exact joint likelihood of the tokenized action sequence: action dimensions can differ in entropy, calibration, and task/timestep relevance, and isolated low-probability tokens may not indicate genuine execution risk.

\begin{table}[!t]
    \centering
    \caption{Comparison of different strategies for aggregating confidence across the set of tokens produced by the VLA models.  Results are presented (1) per model, averaged across task suites and (2) averaged across all models and task suites.  Using the arithmetic mean performs best on average for $ECE_1$ and the proper scoring rules, but the geometric mean offers comparable performance. }
    \begin{tabular}{llcccc}
    \toprule
    Model & Method & $ECE_1$ & $ECE_2$ & Brier score & NLL \\
     \midrule
    NORA & Arithmetic Mean & 0.140 & 0.158 & 0.142 & 0.483 \\
     & Geometric Mean & 0.162 & 0.185 & 0.150 & 0.501 \\
     & Max & 0.148 & 0.152 & 0.148 & 2.006 \\
     & Min & 0.442 & 0.468 & 0.354 & 0.936 \\
    \midrule
    MolmoAct & Arithmetic Mean & 0.116 & 0.127 & 0.137 & 0.573 \\
     & Geometric Mean & 0.113 & 0.124 & 0.136 & 0.568 \\
     & Max & 0.147 & 0.156 & 0.147 & 2.000 \\
     & Min & 0.083 & 0.097 & 0.128 & 0.460 \\
     \midrule
    UniVLA & Arithmetic Mean & 0.123 & 0.166 & 0.069 & 0.276 \\
     & Geometric Mean & 0.144 & 0.189 & 0.078 & 0.298 \\
     & Max & 0.049 & 0.059 & 0.046 & 0.589 \\
     & Min & 0.336 & 0.383 & 0.192 & 0.564 \\
     \midrule
    OpenVLA & Arithmetic Mean & 0.170 & 0.185 & 0.216 & 0.756 \\
     & Geometric Mean & 0.160 & 0.175 & 0.211 & 0.732 \\
     & Max & 0.251 & 0.251 & 0.251 & 3.446 \\
     & Min & 0.220 & 0.266 & 0.239 & 0.699 \\
     \midrule
    Average & Arithmetic Mean & 0.137 & 0.159 & 0.141 & 0.522 \\
     & Geometric Mean & 0.145 & 0.168 & 0.144 & 0.525 \\
     & Max & 0.148 & 0.155 & 0.148 & 2.010 \\
     & Min & 0.270 & 0.303 & 0.228 & 0.665 \\
    \bottomrule
    \end{tabular}
    \label{tab:aggregation_comparison}
\end{table}

\subsection{Relationship Between Task Success and Calibration}


Table \ref{tab:all_results} features all results plotted in Figure \ref{fig:tradeoff_baseline}.

\begin{table*}[t]
    \centering
    \caption{All task success and calibration results for 22 VLA/task suite combinations.}
    \begin{tabular}{cccccccc}
    \toprule
    Dataset & Model & Quant & $\text{ECE}_1$ & $\text{ECE}_2$ & Brier & NLL & Succ. Rate \\
    \midrule
    Spatial & NORA & - &  0.099 & 0.115 & 0.105 & 0.386 & 0.894 \\
    Object & NORA & - &  0.210 & 0.229 & 0.090 & 0.343 & 0.960 \\
    Goal & NORA & - &  0.065 & 0.077 & 0.111 & 0.391 & 0.880 \\
    10 & NORA & - &  0.184 & 0.212 & 0.260 & 0.811 & 0.674 \\
    \midrule
    Spatial & MolmoAct & - &  0.113 & 0.123 & 0.134 & 0.625 & 0.860 \\
    Object & MolmoAct & - &  0.046 & 0.059 & 0.062 & 0.296 & 0.936 \\
    Goal & MolmoAct & - &  0.129 & 0.141 & 0.146 & 0.657 & 0.846 \\
    10 & MolmoAct & - &  0.175 & 0.184 & 0.205 & 0.716 & 0.772 \\
    \midrule
    Spatial & UniVLA & - &  0.162 & 0.193 & 0.064 & 0.282 & 0.972 \\
    Object & UniVLA & - &  0.082 & 0.135 & 0.054 & 0.224 & 0.962 \\
    Goal & UniVLA & - &  0.116 & 0.163 & 0.066 & 0.261 & 0.958 \\
    10 & UniVLA & - &  0.134 & 0.174 & 0.092 & 0.339 & 0.932 \\
    \midrule
    Spatial & OpenVLA & - &  0.088 & 0.106 & 0.150 & 0.533 & 0.828 \\
    Object & OpenVLA & - &  0.060 & 0.073 & 0.108 & 0.401 & 0.880 \\
    Goal & OpenVLA & - &  0.151 & 0.170 & 0.207 & 0.707 & 0.758 \\
    10 & OpenVLA & - &  0.381 & 0.390 & 0.398 & 1.382 & 0.532 \\
    \midrule
    Spatial & OpenVLA & Quant-8 &  0.070 & 0.093 & 0.138 & 0.488 & 0.844 \\
    Object & OpenVLA & Quant-8 &  0.057 & 0.065 & 0.110 & 0.409 & 0.876 \\
    Goal & OpenVLA & Quant-8 &  0.140 & 0.163 & 0.205 & 0.713 & 0.760 \\
    \midrule
    Spatial & OpenVLA & Quant-4 &  0.067 & 0.086 & 0.149 & 0.503 & 0.822 \\
    Object & OpenVLA & Quant-4 &  0.091 & 0.112 & 0.135 & 0.489 & 0.850 \\
    Goal & OpenVLA & Quant-4 &  0.161 & 0.182 & 0.231 & 0.748 & 0.712 \\
    \bottomrule
    \end{tabular}
    \label{tab:all_results}
\end{table*}

\subsubsection{Controlling for marginal success-rate effects}
\label{app:marginal_effects}

\begin{figure*}[t]
\centering
\includegraphics[width=0.8\textwidth]{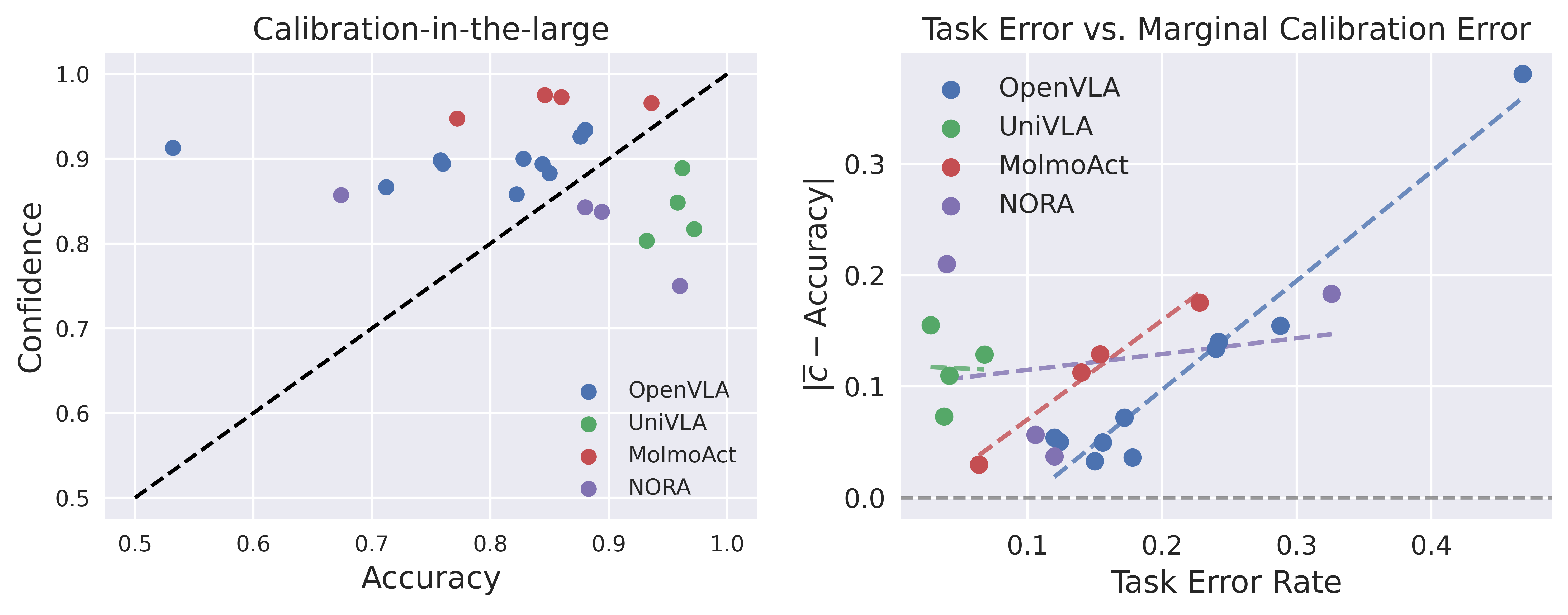}
\caption{Calibration-in-the-large across model--task suite pairs. Left: mean first-step confidence versus empirical success rate, with the dashed line denoting perfect marginal calibration. Right: marginal calibration error $|\bar{c}-\bar{y}|$ versus task error, with model-family trend lines.}
\label{fig:cal_in_the_large}
\end{figure*}

To determine how much of the relationship between task performance and calibration is explained by marginal confidence--success alignment, we first examine calibration-in-the-large. For each model--task suite pair, we compute the mean first-step confidence $\bar{c}$ and empirical success rate $\bar{y}$. As shown in Figure~\ref{fig:cal_in_the_large}, OpenVLA and MolmoAct generally move closer to the $\bar{c}=\bar{y}$ line as task error decreases, whereas UniVLA and NORA exhibit weaker or less consistent within-family trends. Calibration-in-the-large, measured by $|\bar{c}-\bar{y}|$, is equivalent here to ECE computed with a single bin. It is therefore a useful diagnostic of marginal alignment, although collapsing all predictions into one bin can conceal miscalibration across different confidence levels. Overall, improved alignment between mean confidence and mean success explains part of the lower calibration error observed in higher-performing settings, but the strength of this relationship differs across VLA families.

\begin{figure*}[t]
\centering
\includegraphics[width=\textwidth]{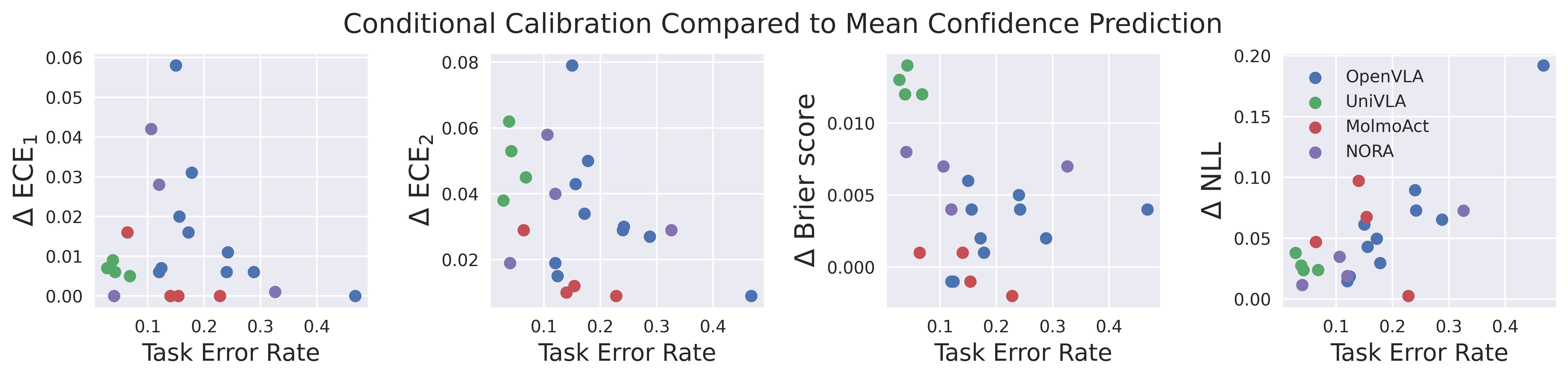}
\caption{Calibration metrics relative to the constant mean-confidence control. We report $\Delta M=M(\mathbf{c},\mathbf{y})-M(\bar{c},\mathbf{y})$; positive values indicate higher error for the original trial-level confidence estimates.}
\label{fig:delta_metrics}
\end{figure*}

\begin{figure*}[t]
\centering
\includegraphics[width=0.85\textwidth]{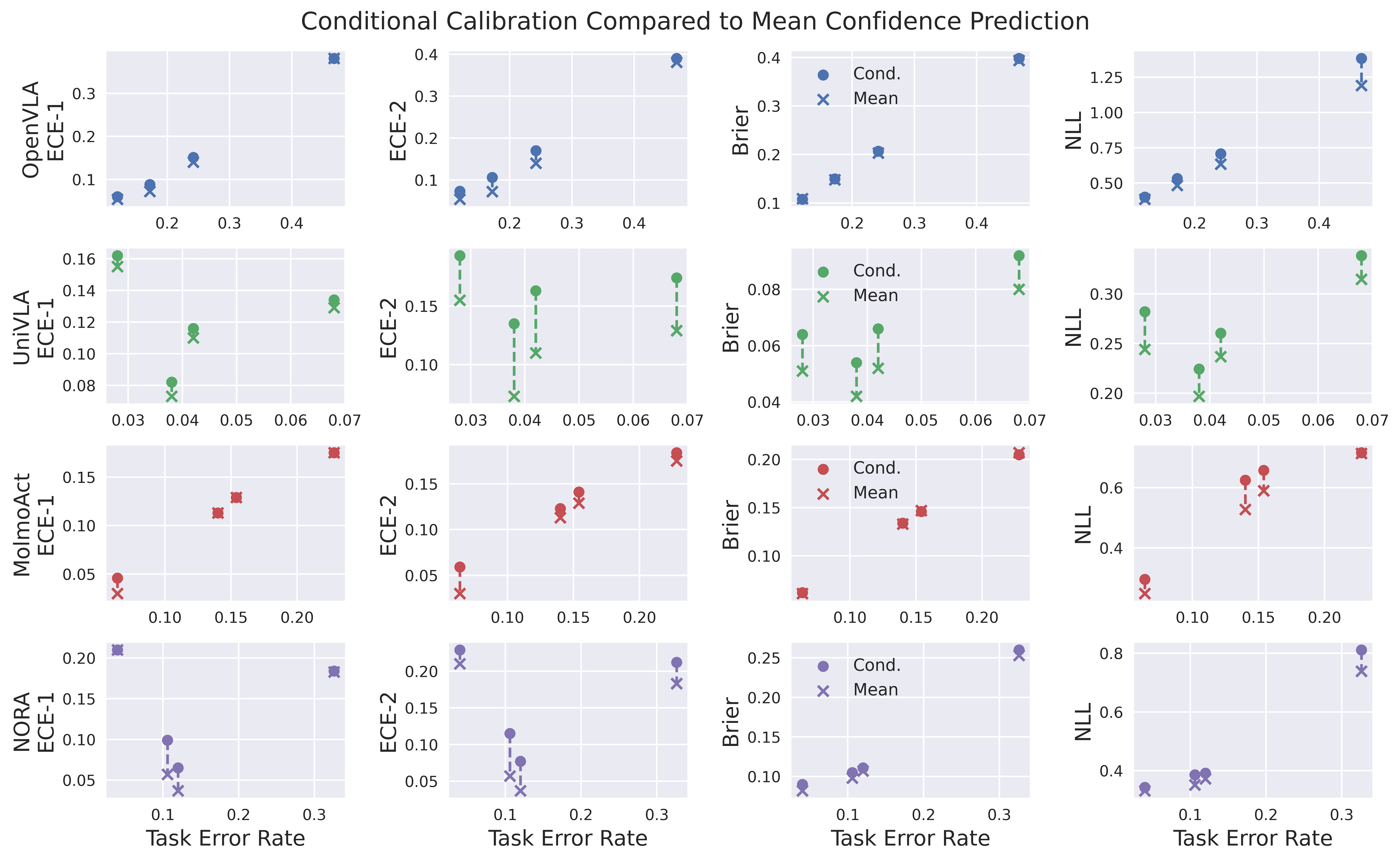}
\caption{Calibration metrics for the original trial-level confidence estimates and the corresponding constant mean-confidence controls across model families and task suites.}
\label{fig:cond_vs_mean_cal}
\end{figure*}

We next isolate the contribution of variation in confidence across individual trials by constructing a constant predictor for each model--task suite pair. Specifically, we replace every trial-level confidence $c_i$ with the corresponding pairwise mean $\bar{c}$. This control preserves both the mean confidence and the task-suite success rate while removing all within-suite variation. Figure~\ref{fig:cond_vs_mean_cal} directly compares the resulting ECE$_1$, ECE$_2$, Brier score, and NLL values with those obtained from the original trial-level confidence estimates. Figure~\ref{fig:delta_metrics} summarizes the same comparison using
\[
\Delta M
=
M(\mathbf{c},\mathbf{y})
-
M(\bar{c},\mathbf{y}),
\]
where positive values indicate greater error for the original trial-varying predictor than for the constant control.

Because the constant predictor emits only one confidence value, its ECE is precisely the marginal calibration error $|\bar{c}-\bar{y}|$. Consequently, the ECE comparison primarily quantifies the additional calibration structure revealed by using multiple confidence levels; we rely especially on Brier score and NLL to determine whether the observed relationship extends beyond calibration-in-the-large. The results show that marginal alignment accounts for a meaningful portion of the cross-setting variation, while architecture- and metric-dependent differences remain. Differences in Brier score are generally small, whereas NLL reveals larger differences in several settings, indicating that trial-level variation can be particularly sensitive to highly confident failures.

\subsection{Prompt Ensembling}

Table \ref{tab:appendix_calibration} has prompt ensemble results for all models.

\begin{table*}[t]
    \centering
    \caption{Calibration error measurements using all 4 models (and the OpenVLA 8-bit and 4-bit fine-tuned model versions) and 3 different LIBERO task suites, for 2 different methods of confidence estimation: (1) baseline (2) ensembling over semantically equivalent prompts (``Reprompt'').  The prompt ensemble method improves most measures. }
    \begin{tabular}{lllcccc}
    \toprule
    Model & Dataset & Method & $\text{ECE}_1$ & $\text{ECE}_2$ & Brier score & NLL \\
    \midrule
    OpenVLA & Spatial & Baseline & 0.0879 & 0.1062 & 0.1498 & 0.5330 \\
    ~ & ~ & Reprompt & 0.0525 & 0.0685 & 0.1449 & 0.4766 \\
    ~ & Object & Baseline & 0.0602 & 0.0727 & 0.1076 & 0.4010 \\
    ~ & ~ & Reprompt & 0.0363 & 0.0527 & 0.1046 & 0.3610 \\
    ~ & Goal & Baseline & 0.1513 & 0.1701 & 0.2065 & 0.7072 \\
    ~ & ~ & Reprompt & 0.1155 & 0.1322 & 0.1974 & 0.6195 \\
    \midrule
    OpenVLA (Quant 8) & Spatial & Baseline & 0.0704 & 0.0934 & 0.1378 & 0.4876 \\
    ~ & ~ & Reprompt & 0.0498 & 0.0616 & 0.1311 & 0.4342 \\
    ~ & Object & Baseline & 0.0574 & 0.0651 & 0.1096 & 0.4089 \\
    ~ & ~ & Reprompt & 0.0414 & 0.0522 & 0.1081 & 0.3745 \\
    ~ & Goal & Baseline & 0.1396 & 0.1627 & 0.2052 & 0.7134 \\
    ~ & ~ & Reprompt & 0.1170 & 0.1520 & 0.1938 & 0.6081 \\
    \midrule
    OpenVLA (Quant 4) & Spatial & Baseline & 0.0673 & 0.0861 & 0.1493 & 0.5029 \\
    ~ & ~ & Reprompt & 0.0546 & 0.0790 & 0.1482 & 0.4825 \\
    ~ & Object & Baseline & 0.0911 & 0.1125 & 0.1354 & 0.4889  \\
    ~ & ~ & Reprompt & 0.0663 & 0.0942 & 0.1318 & 0.4540  \\
    ~ & Goal & Baseline & 0.1615 & 0.1821 & 0.2307 & 0.7475 \\
    ~ & ~ & Reprompt & 0.1456 & 0.1535 & 0.2239 & 0.6720  \\
    \midrule
    NORA & Spatial & Baseline & 0.0986 & 0.1153 & 0.1054 & 0.3859 \\
    ~ & ~ & Reprompt & 0.0918 & 0.1117 & 0.1046 & 0.3758 \\
    UniVLA & Spatial & Baseline & 0.1618 & 0.1933 & 0.0644 & 0.2821 \\
    ~ & ~ & Reprompt & 0.1573 & 0.1924 & 0.0638 & 0.2624 \\
    MolmoAct & Spatial & Baseline & 0.1126 & 0.1231 & 0.1337 & 0.6248 \\
    ~ & ~ & Reprompt & 0.1137 & 0.1199 & 0.1338 & 0.5834 \\
    \bottomrule
    \end{tabular}
    \label{tab:appendix_calibration}
\end{table*}


\begin{figure*}[t]
\centering
\includegraphics[width=\textwidth]{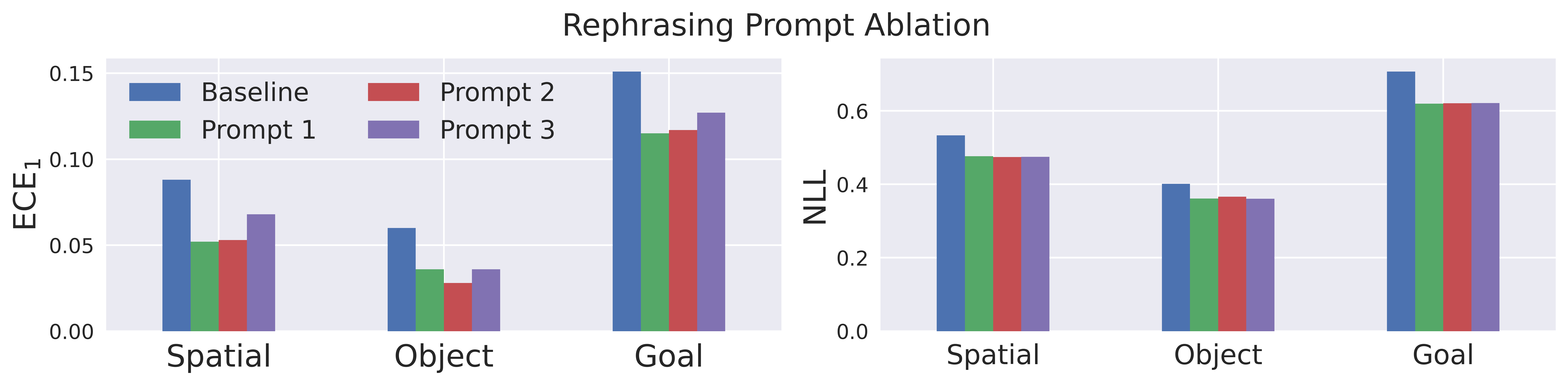}
\caption{Ablation results for the prompt ensemble method, where different prompts are given to GPT-4o-mini for producing the rephrasings of the original instruction.  Prompt 1 is used in the main experiments, while Prompts 2 and 3 are variants.  Results show that improvements in calibration error from Reprompt are robust to different rephrasing prompts.}
\label{fig:rephrasing_ablation}
\end{figure*}

\begin{figure*}[t]
\centering
\includegraphics[width=\textwidth]{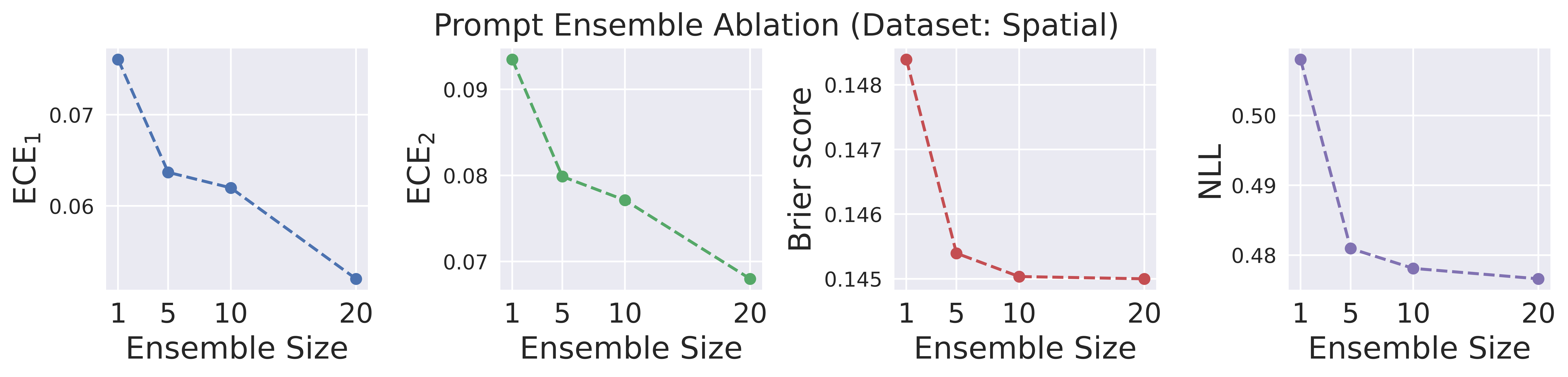}
\includegraphics[width=\textwidth]{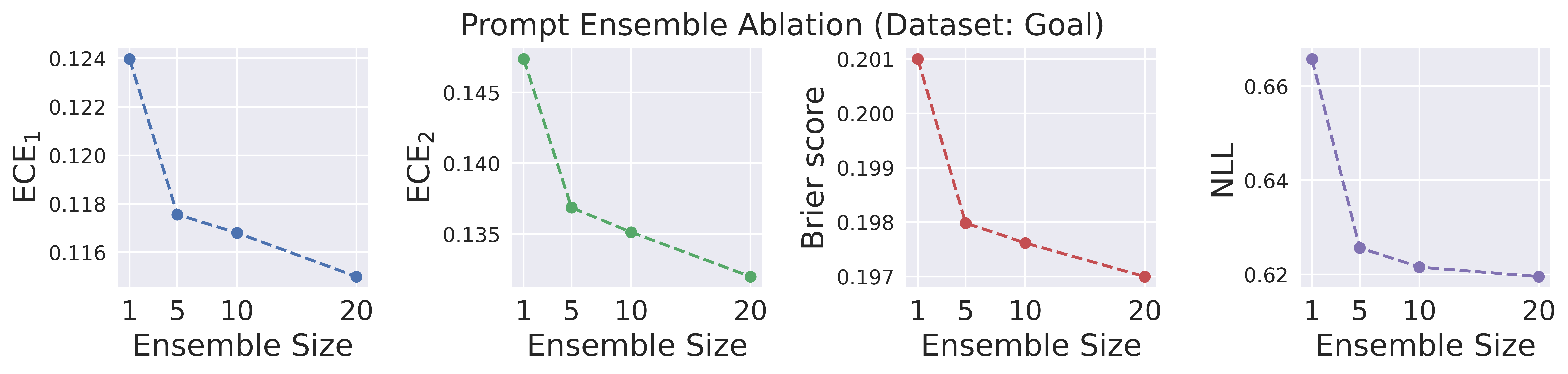}
\includegraphics[width=\textwidth]{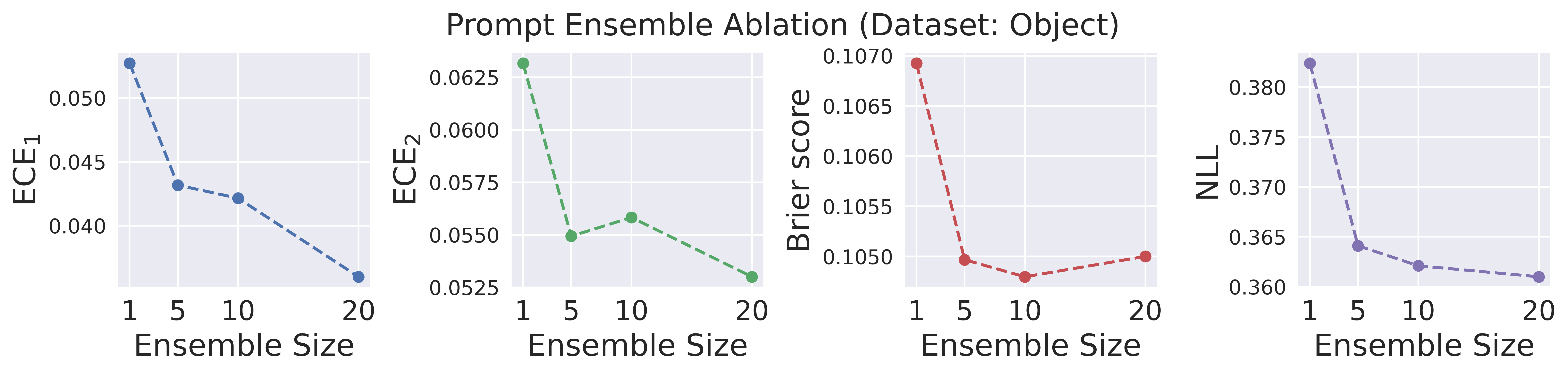}
\caption{Ablating the number of prompts in the prompt ensemble for the Spatial, Goal, and Object task suites.  Increasing the number of prompts generally improves calibration error.}
\label{fig:n_prompt_ablation}
\end{figure*}

\subsection{Ablations for Prompt Ensembles}\label{sec:ablations}
To understand the robustness of our prompt ensemble results, we perform multiple ablations.  
First, we consider the effect of changing the prompt given to GPT-4o-mini for producing the 20 instruction rephrasings.  
In addition to the prompt used to produce the results in Table~\ref{tab:appendix_calibration}, we try two additional prompts, recorded in Appendix Table~\ref{tab:reprompt_prompts}; ``Prompt 1'' is the original prompt, and ``Prompt 2'' and ``Prompt 3'' represent new prompts for the ablation.  
Note that the different prompts lead to substantially different rephrasings, as Prompts 2 and 3 mandate the retention of the words ``pick'' and ``place'', while Table~\ref{tab:reprompting_example} shows that Prompt 1 leads to these words being replaced.
The ablation is run across all 3 task suites using fine-tuned OpenVLA, and we measure $\text{ECE}_1$ and NLL.
Results are shown in Figure~\ref{fig:rephrasing_ablation}.
Improvements in calibration error from the Reprompt method are robust to different rephrasing prompts, with all ensemble variants producing lower error than the baseline method across both metrics and all 3 task suites.

Second, we consider the effect of the number of prompts used in the prompt ensemble.  
Given a sound ensembling approach, we would expect to see improvement as more prompts are added to the ensemble (up to some point).  
To study this question, we record results using $k\in\{1,5,10,20\}$ different rephrasings (randomly chosen for 1000 trials when $k<20$).  
Results for the Spatial, Goal and Object task suites are in Figure~\ref{fig:n_prompt_ablation}.
We see that the algorithm behaves favorably, where calibration error generally decreases as more instructions are included in the ensemble.

\subsection{Further Analysis of Prompt Ensembles}\label{app:reprompt_action_agreement}

In the prompt ensemble experiments, the original instruction determines every
action executed in the environment and therefore determines the binary
episode-success label. At the initial state, we additionally
query the VLA with 20 rephrasings to obtain auxiliary
confidence estimates. The candidate actions produced under the rephrasings are not executed. Thus, the ensemble
confidence is evaluated against the outcome of a fixed
original-instruction policy, rather than against an averaged or voted action.
To characterize how much rephrasing changes the model’s first action, we compare the actions generated under each rephrasing and the original instruction at the same observation, using OpenVLA on the Spatial, Object, and Goal task suites.
Before computing agreement, we apply the same OpenVLA
normalization, binarization, and sign convention to the gripper coordinate
used for environment execution. 
Results for the three OpenVLA task suites in
the top row of Figure~\ref{fig:reprompt_reductions} are reported in
Table~\ref{tab:reprompt_action_agreement}.

\begin{table}[t]
\centering
\caption{
Agreement between first actions produced by OpenVLA under the original
instruction and 20 semantically equivalent rephrasings. Dimension agreement
averages exact agreement across the six non-gripper motion dimensions; exact
6-D agreement requires all six motion coordinates to match. ``Within one bin''
allows each motion coordinate to differ by at most one OpenVLA token bin.
MAE is measured for the 6 motion dimensions on the normalized $[-1,1]$ action scale.
}
\begin{tabular}{lccccc}
\toprule
Task
& Dimension agreement
& Exact 6-D agreement
& Within one bin
& Gripper agreement
& MAE \\
\midrule
Spatial & 49.56\% & 13.40\% & 16.30\% & 99.85\% & 0.0453 \\
Object  & 65.20\% & 42.70\% & 45.66\% & 99.18\% & 0.0216 \\
Goal    & 51.76\% & 10.38\% & 13.32\% & 99.98\% & 0.0384 \\
\bottomrule
\end{tabular}
\label{tab:reprompt_action_agreement}
\end{table}

Rephrasings do not always produce identical motion action tokens.
Averaged across the six motion dimensions, exact agreement ranges from
49.56\% to 65.20\%, while exact agreement across the complete six-dimensional
action ranges from 10.38\% to 42.70\%. The complete-action criterion is
strict: a difference in any single discretized coordinate makes the action
count as different. Gripper decisions are nearly invariant, with agreement
above 99\% in every suite.
The motion differences are relatively limited in aggregate.
Mean absolute differences range from 0.0216 to 0.0453 on the normalized
$[-1,1]$ action output scale. Moreover, at least one of the 20 rephrasings exactly matches
the original six-dimensional action in 56.60\% of Spatial trials, 84.80\% of
Object trials, and 50.40\% of Goal trials.

\clearpage

\subsection{Calibration Across Task Time}\label{app:task_time}

\begin{figure*}[t]
\centering
\includegraphics[width=\textwidth]{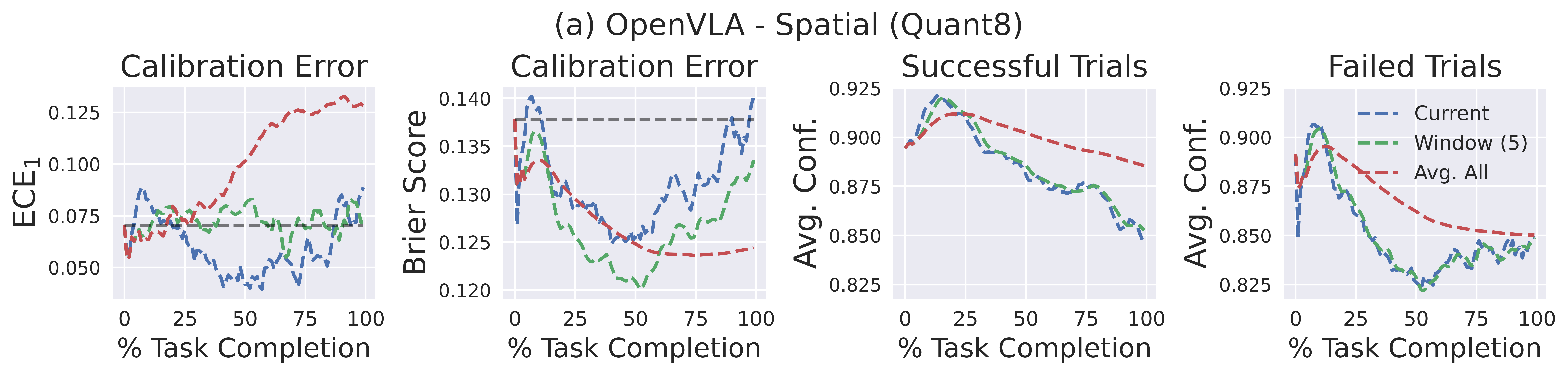}
\includegraphics[width=\textwidth]{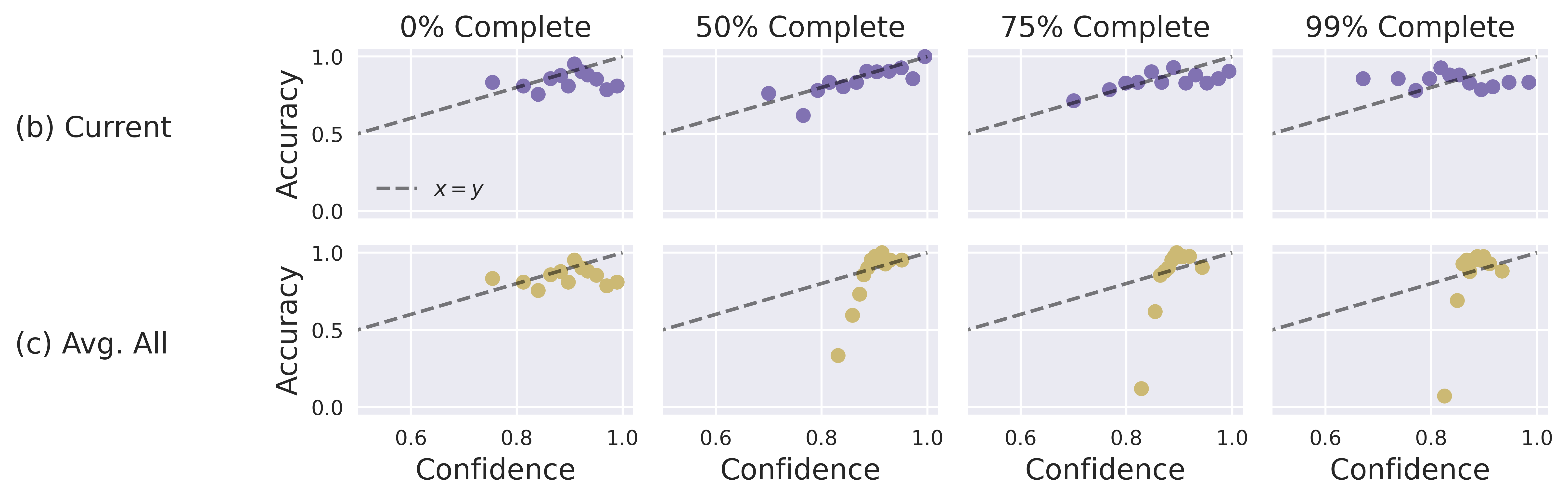}
\caption{Empirical study of calibration error across task time horizon for the Spatial task suite and the Quant-8 OpenVLA model.}
\label{fig:across_time_spatial_quant8}
\end{figure*}

\begin{figure*}[t]
\centering
\includegraphics[width=\textwidth]{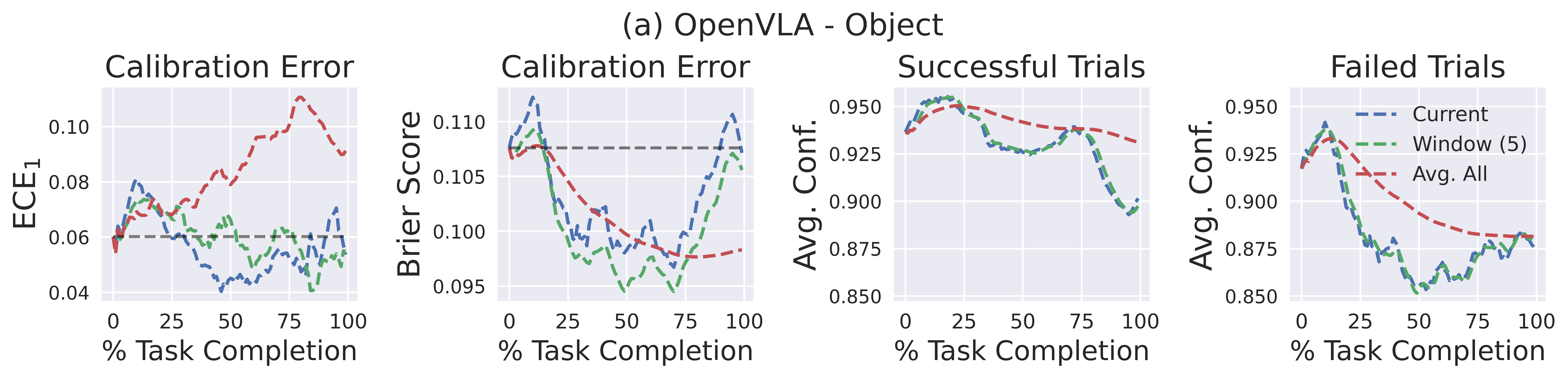}
\includegraphics[width=\textwidth]{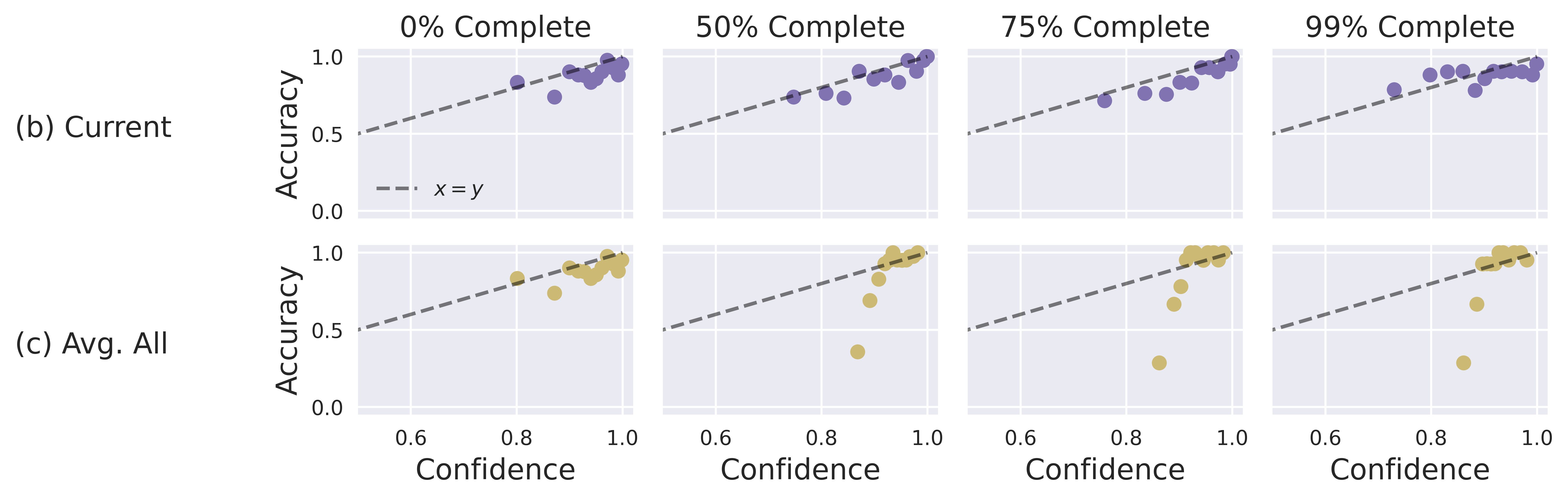}
\caption{Empirical study of calibration error across task time horizon for the Object task suite and the full precision OpenVLA model.}
\label{fig:across_time_object}
\end{figure*}

\begin{figure*}[t]
\centering
\includegraphics[width=\textwidth]{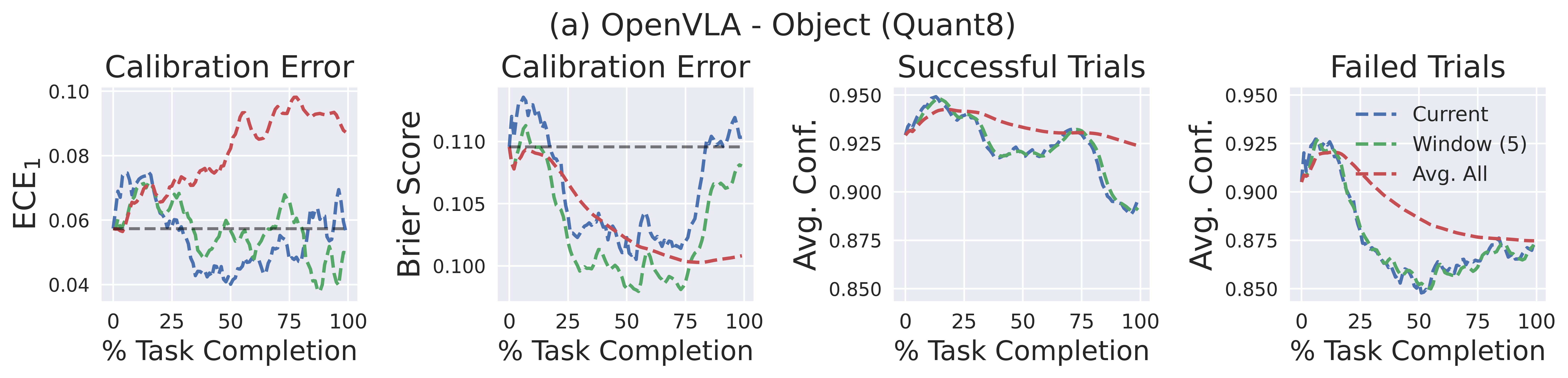}
\includegraphics[width=\textwidth]{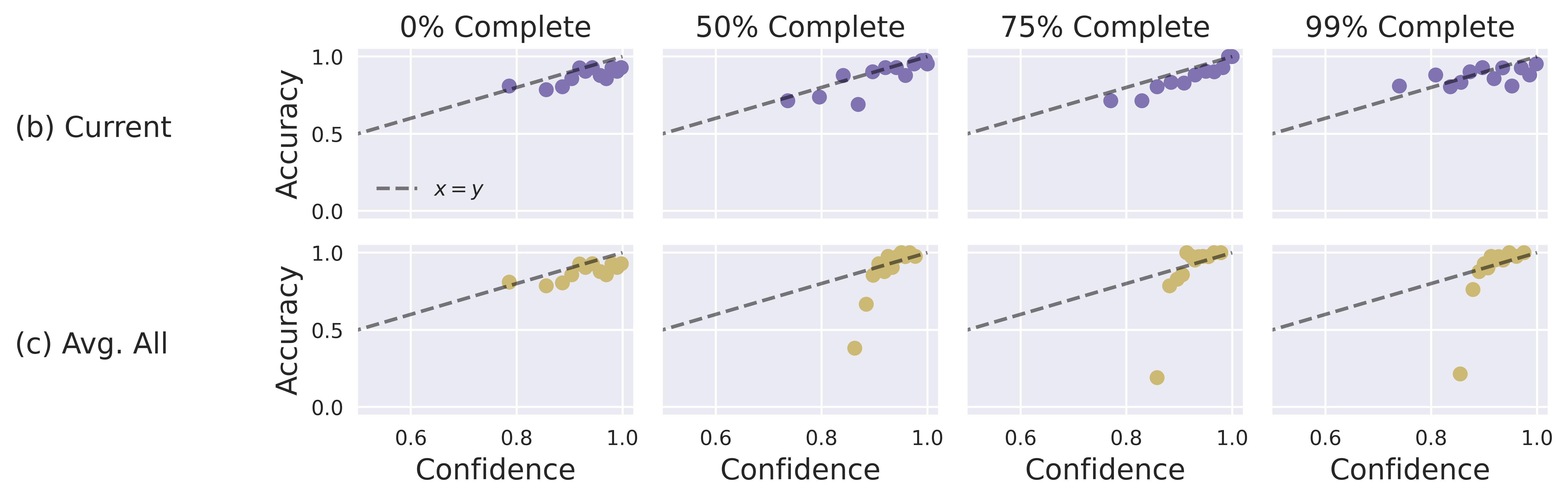}
\caption{Empirical study of calibration error across task time horizon for the Object task suite and the Quant-8 OpenVLA model.}
\label{fig:across_time_object_quant8}
\end{figure*}

\begin{figure*}[t]
\centering
\includegraphics[width=\textwidth]{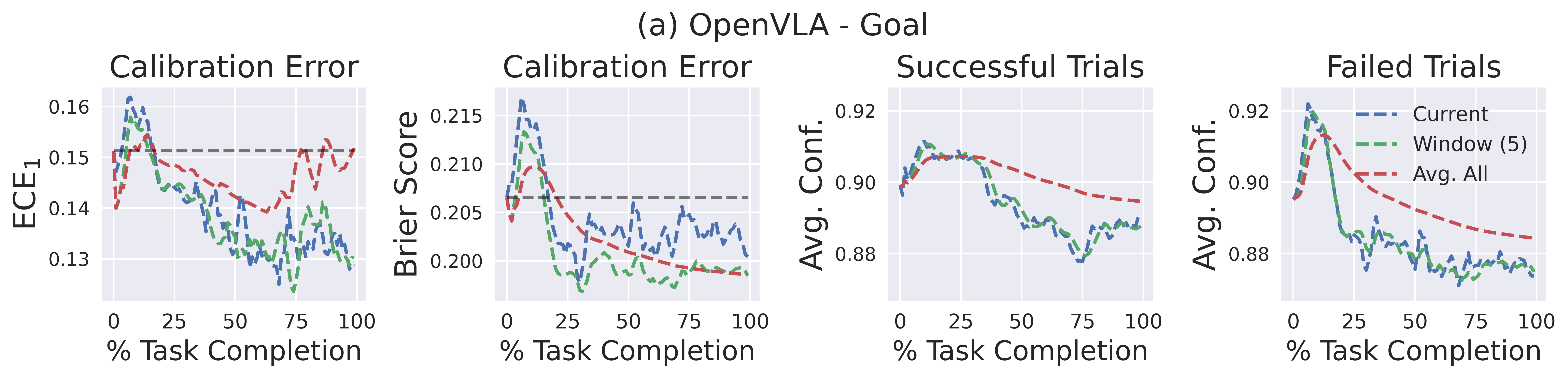}
\includegraphics[width=\textwidth]{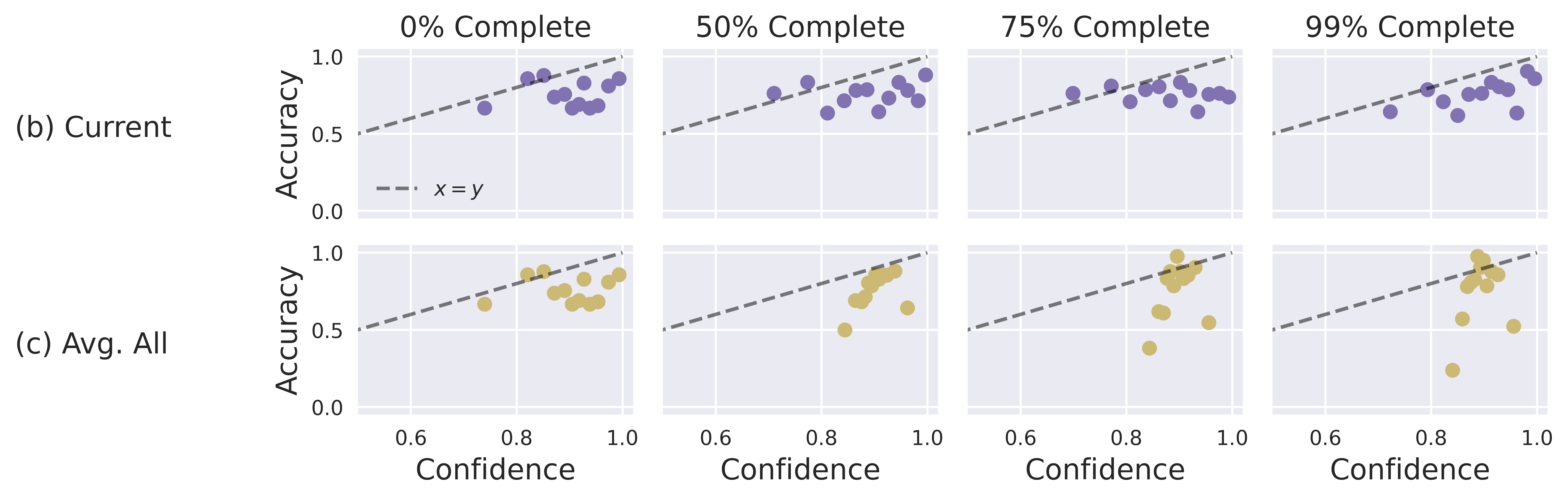}
\caption{Empirical study of calibration error across task time horizon for the Goal task suite and the full precision OpenVLA model.}
\label{fig:across_time_goal}
\end{figure*}

\begin{figure*}[t]
\centering
\includegraphics[width=\textwidth]{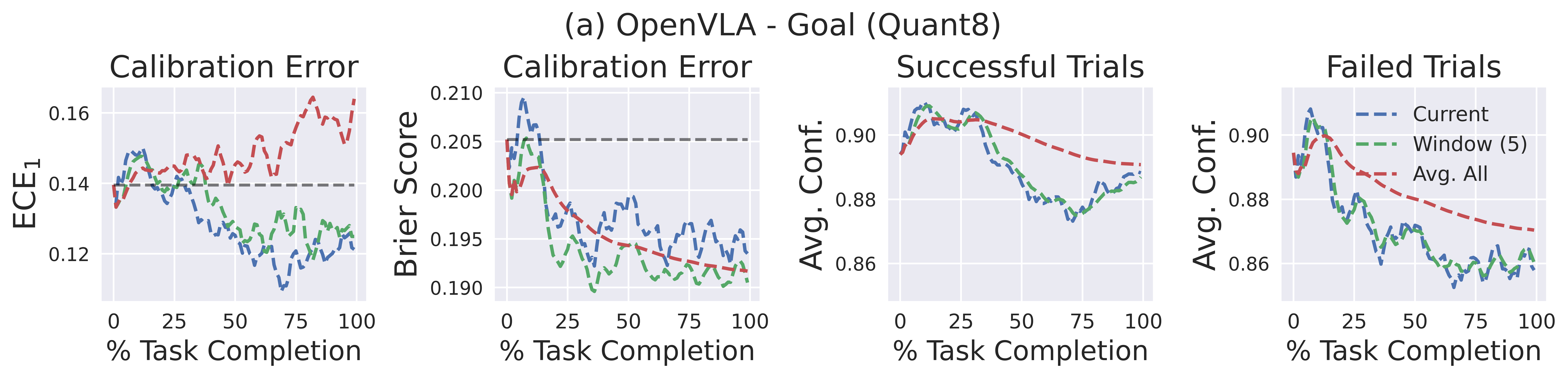}
\includegraphics[width=\textwidth]{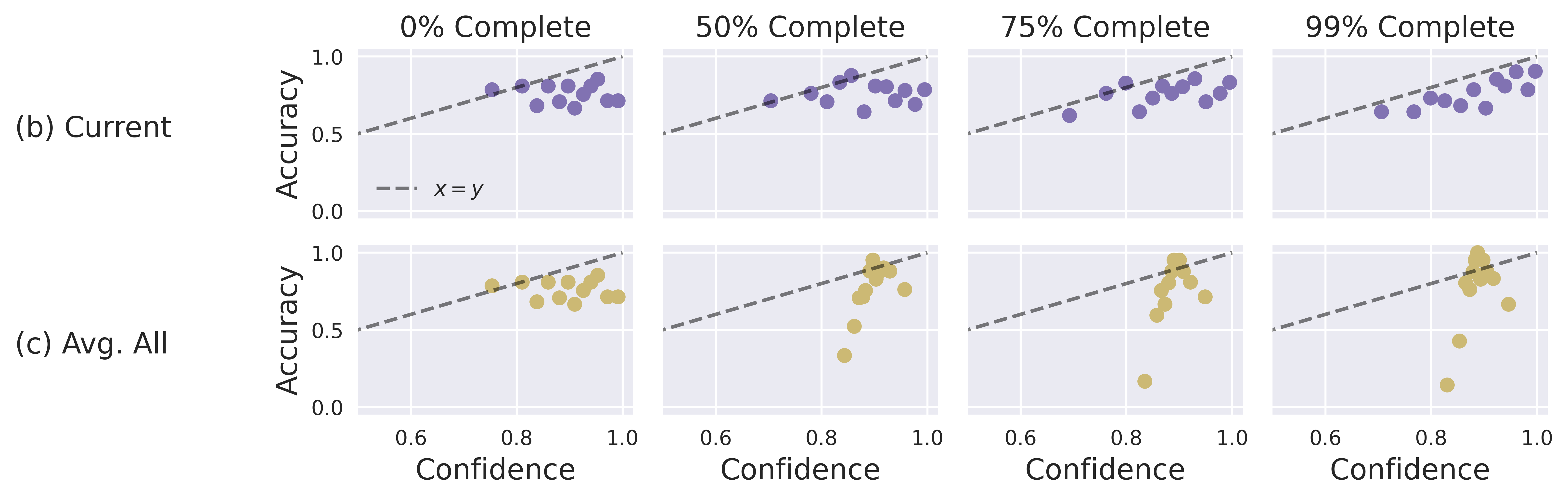}
\caption{Empirical study of calibration error across task time horizon for the Goal task suite and the Quant-8 OpenVLA model.}
\label{fig:across_time_goal_quant8}
\end{figure*}

\begin{figure*}[t]
\centering
\includegraphics[width=\textwidth]{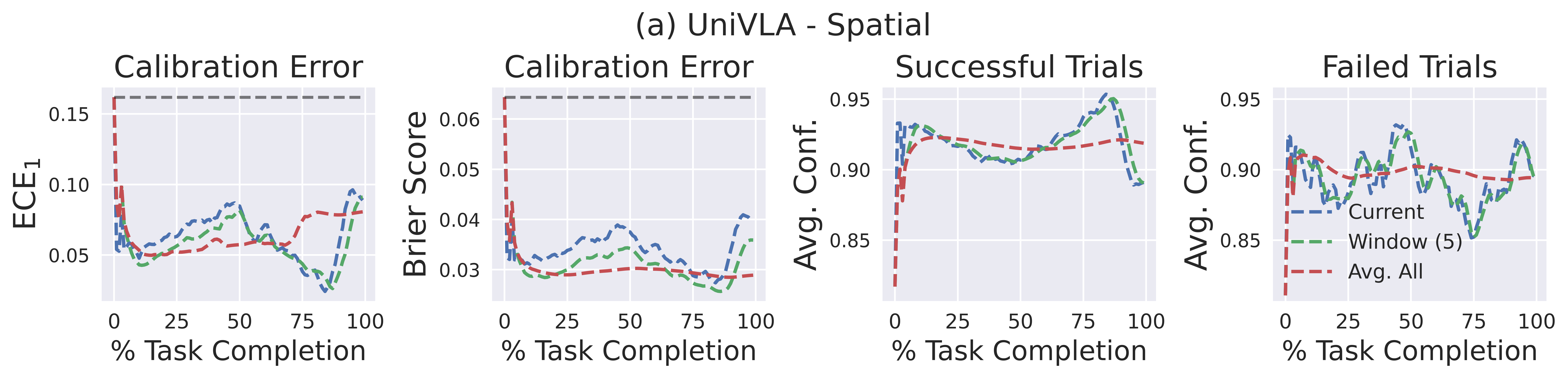}
\includegraphics[width=\textwidth]{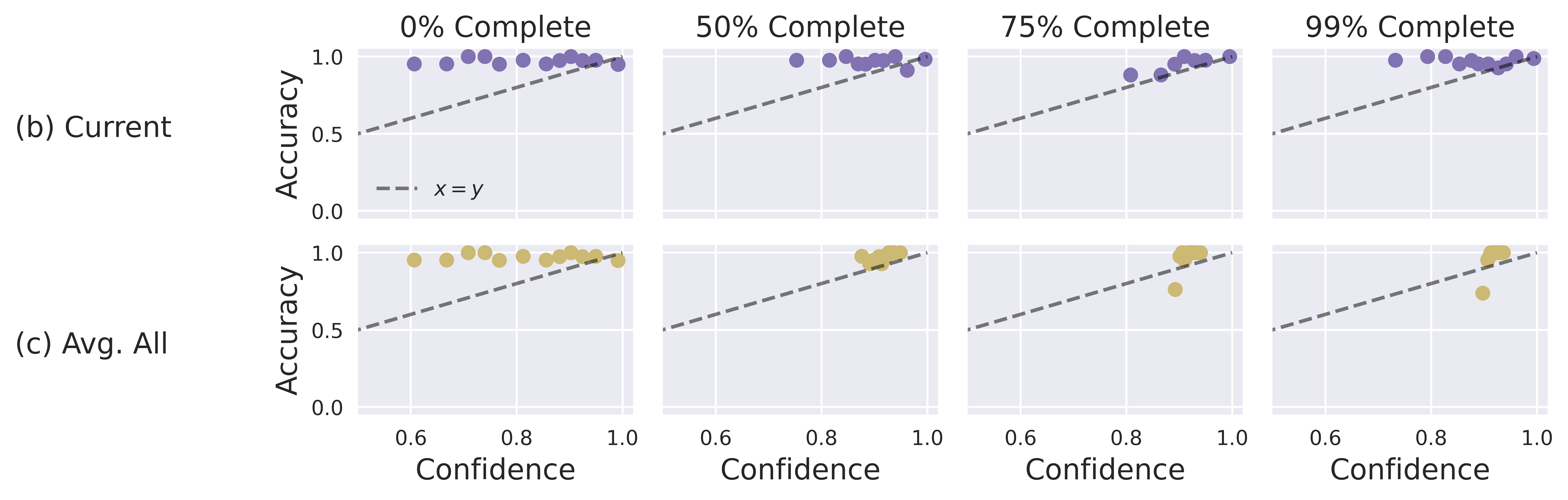}
\caption{Empirical study of calibration error across task time horizon for the Spatial task suite and the UniVLA model.}
\label{fig:across_time_univla_spatial}
\end{figure*}

\begin{figure*}[t]
\centering
\includegraphics[width=\textwidth]{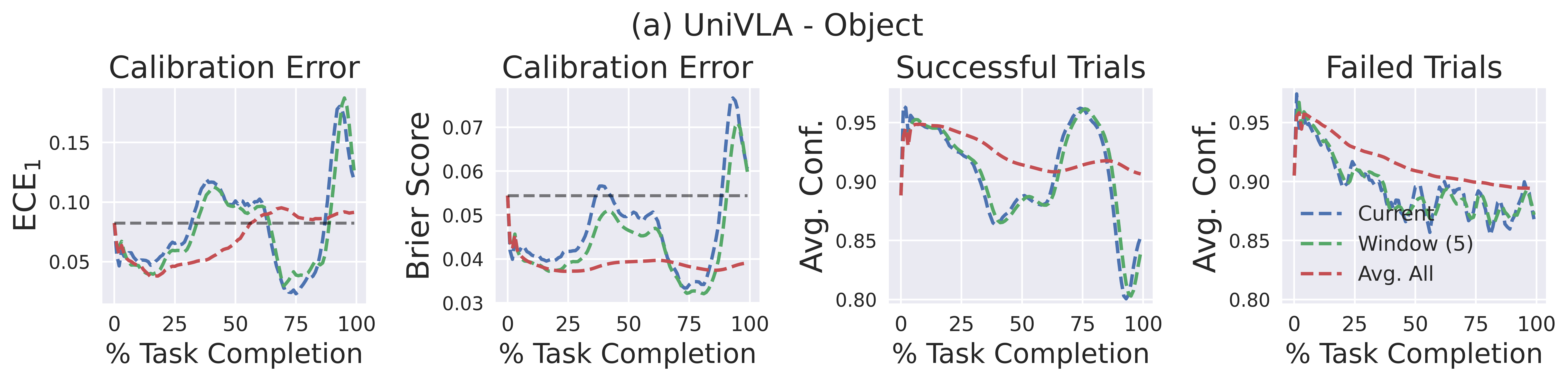}
\includegraphics[width=\textwidth]{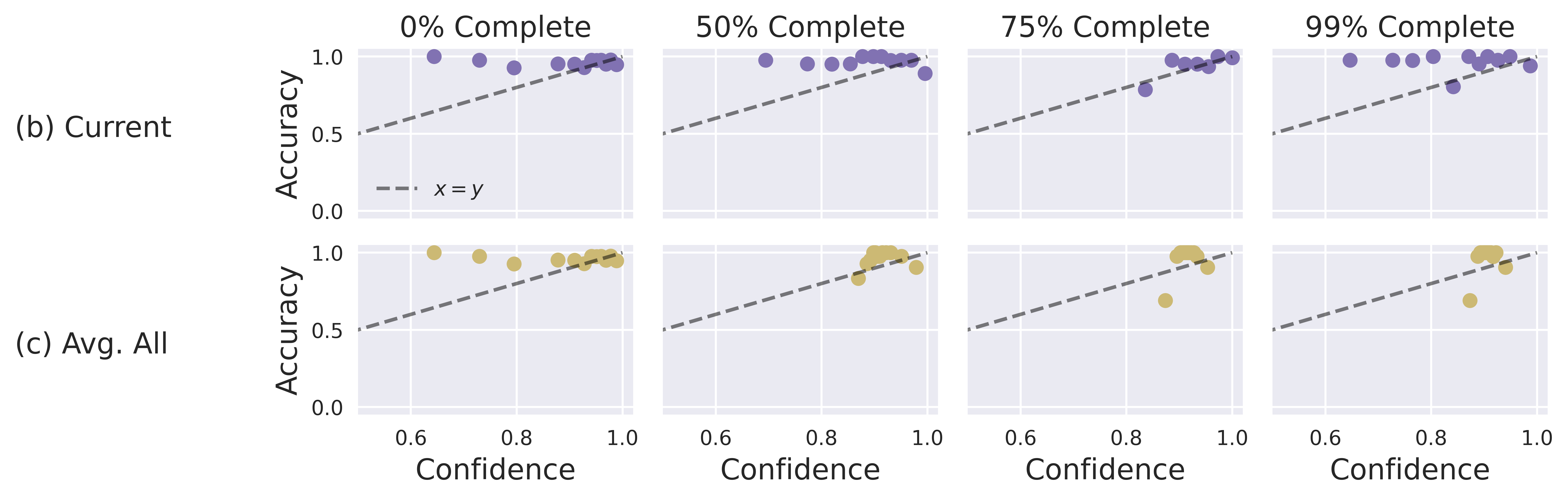}
\caption{Empirical study of calibration error across task time horizon for the Object task suite and the UniVLA model.}
\label{fig:across_time_univla_object}
\end{figure*}

\begin{figure*}[t]
\centering
\includegraphics[width=\textwidth]{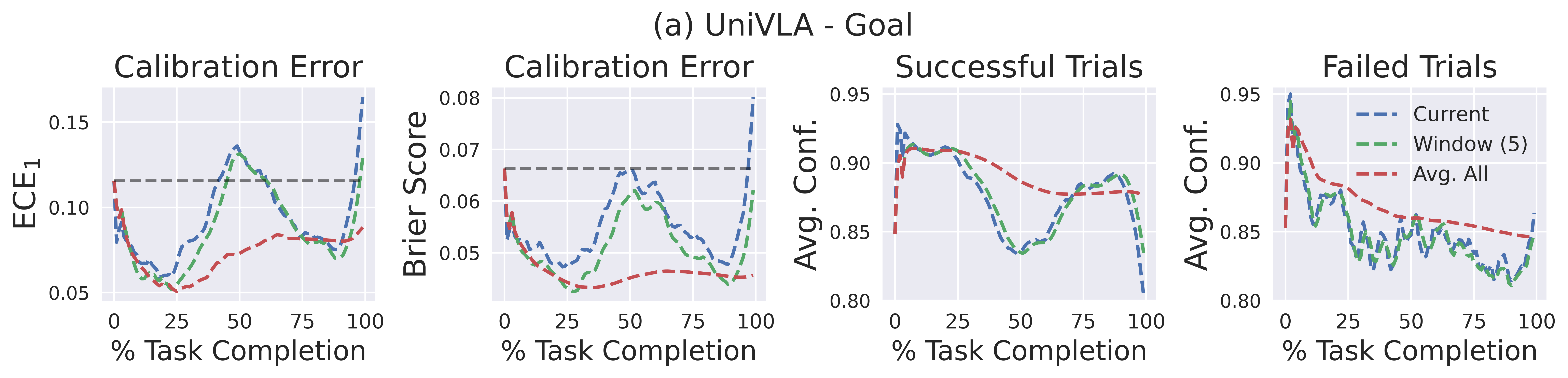}
\includegraphics[width=\textwidth]{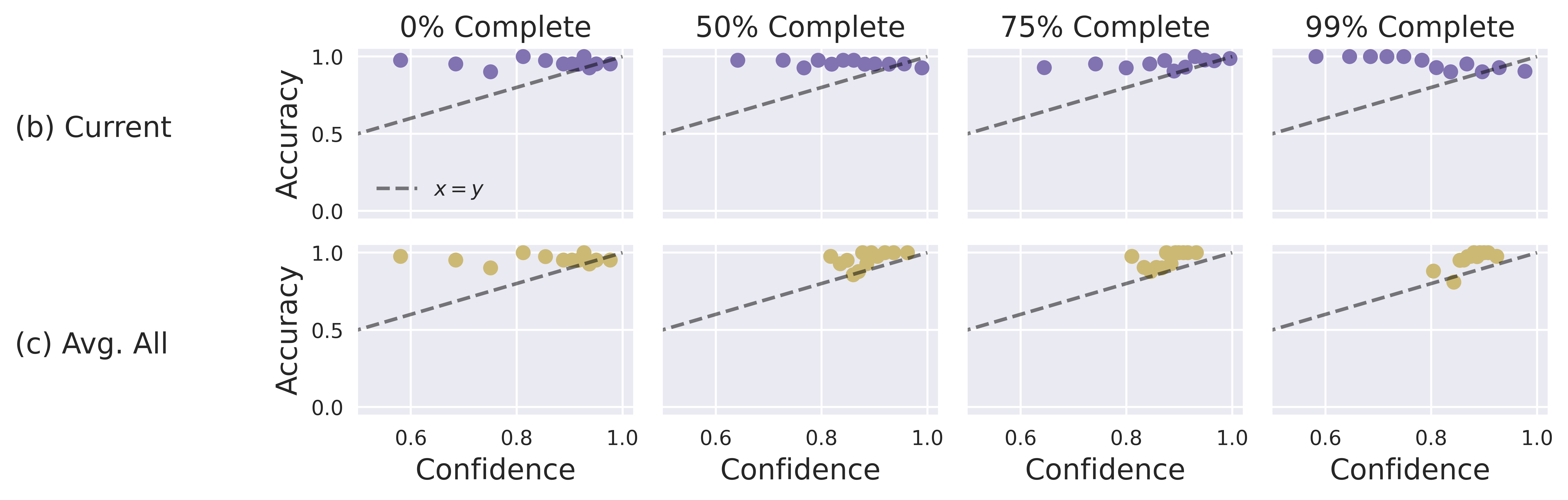}
\caption{Empirical study of calibration error across task time horizon for the Goal task suite and the UniVLA model.}
\label{fig:across_time_univla_goal}
\end{figure*}

Figures \ref{fig:across_time_spatial_quant8}, \ref{fig:across_time_object}, \ref{fig:across_time_object_quant8}, \ref{fig:across_time_goal}, and \ref{fig:across_time_goal_quant8} offer additional results on calibration across the task time horizon using the OpenVLA model.
Figures \ref{fig:across_time_univla_spatial}, \ref{fig:across_time_univla_object}, and \ref{fig:across_time_univla_goal} offer results for the same experiment on UniVLA.
Figure \ref{fig:qualitative_across_time_appendix} shows further qualitative examples from the context-aware monitoring experiment.

\clearpage

\subsubsection{Qualitative Examples of Context-Aware Confidence Monitoring }

\begin{figure*}[t]
\centering
\includegraphics[width=\textwidth]{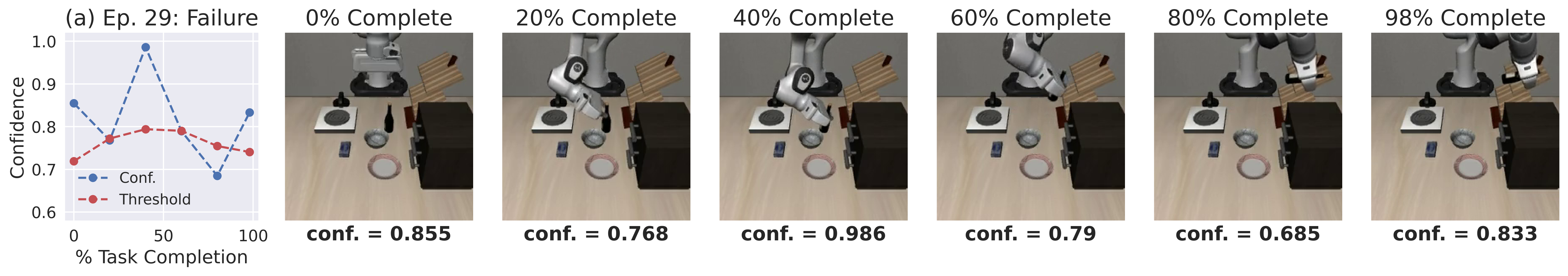}
\includegraphics[width=\textwidth]{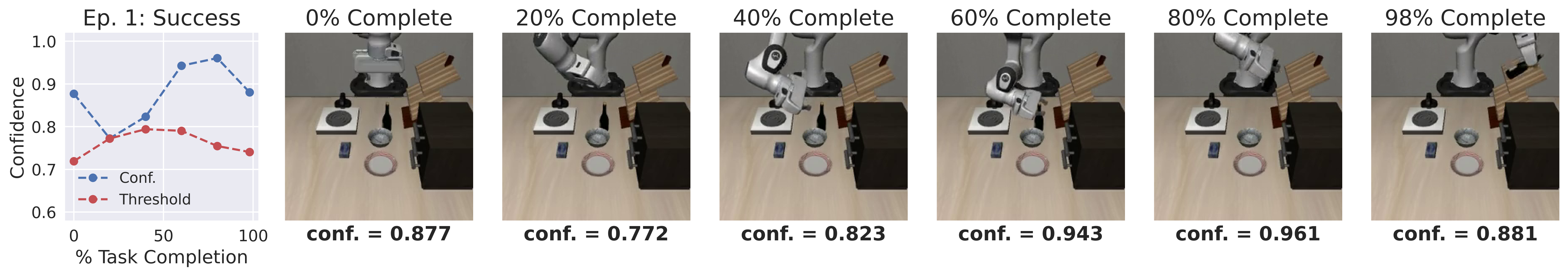}
\includegraphics[width=\textwidth]{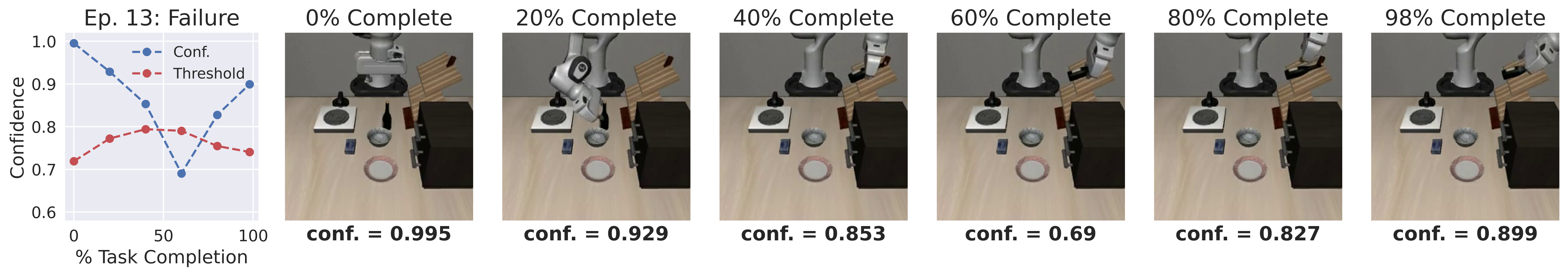}
\includegraphics[width=\textwidth]{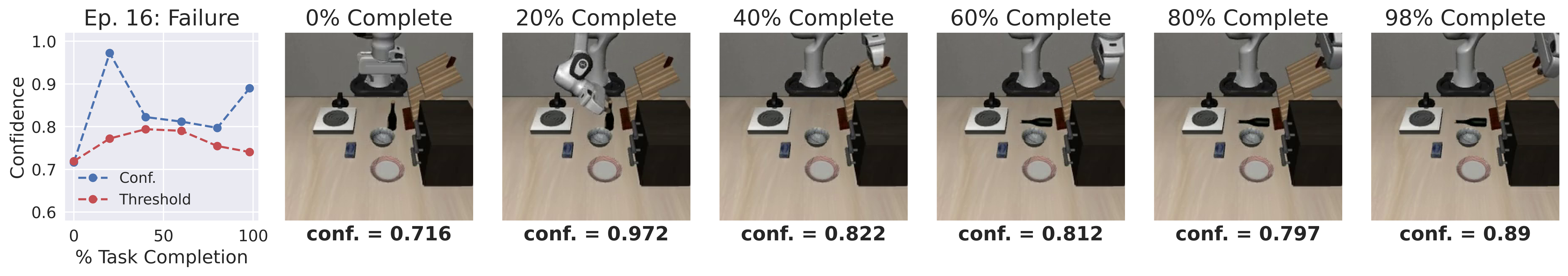}
\includegraphics[width=\textwidth]{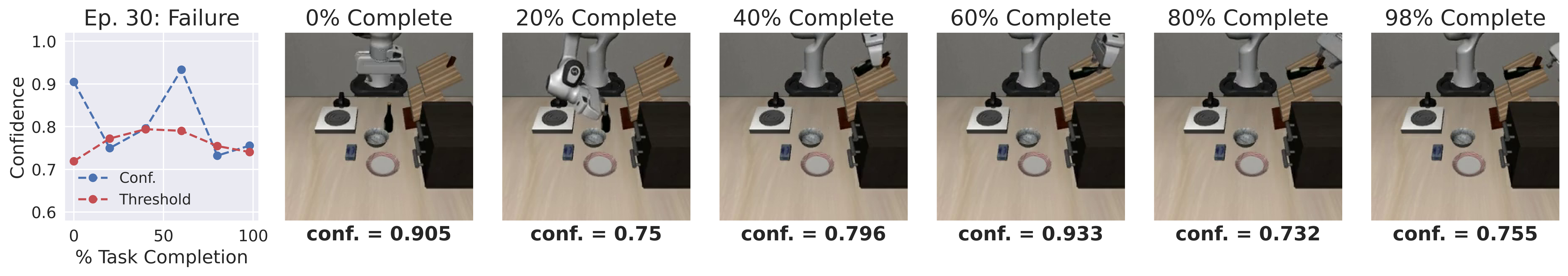}
\includegraphics[width=\textwidth]{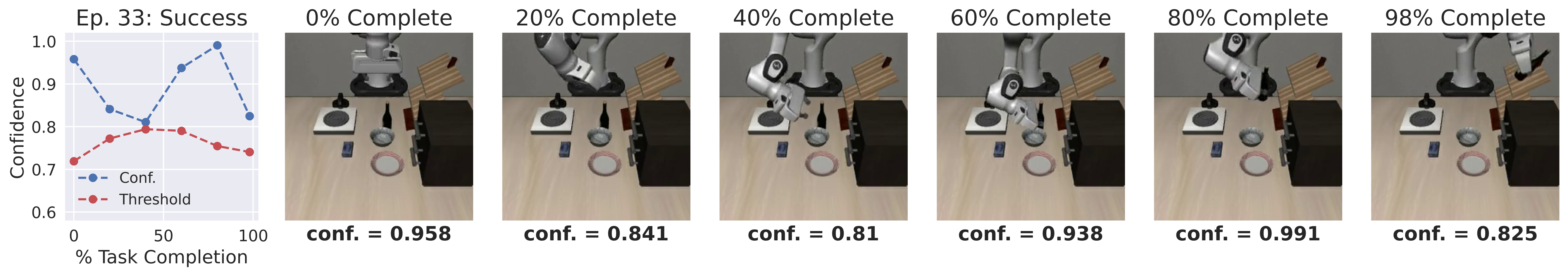}
\includegraphics[width=\textwidth]{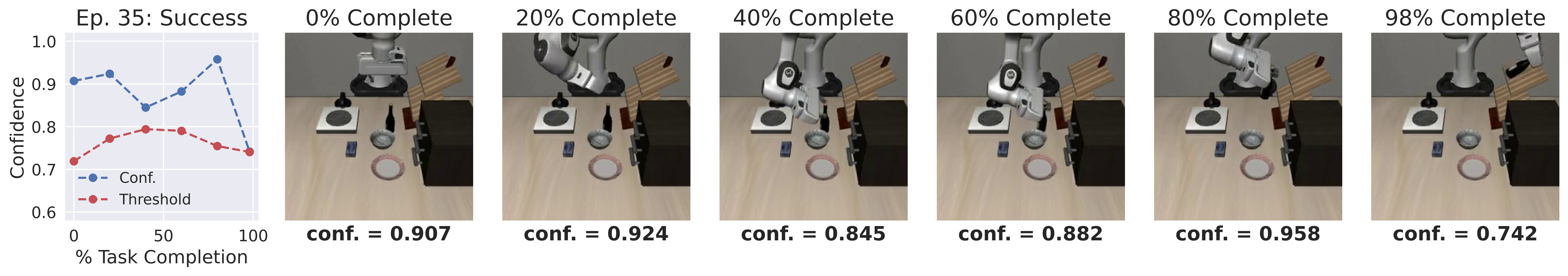}
\caption{Qualitative examples of a context-aware confidence calibration strategy for the task ``put the wine bottle on the rack''.  The red dashed line represents the 10\% quantile of the confidence estimates output by the model across the task horizon, representing a potential threshold below which the robot may abstain from performing the task. }
\label{fig:qualitative_across_time_appendix}
\end{figure*}

Additional qualitative examples of our confidence-aware monitoring demonstration are provided in Figure \ref{fig:qualitative_across_time_appendix}.
These include cases where the strategy succeeds and others where it fails (e.g., confidence falls below the threshold while grasping in an ultimately successful trial).

Consider the episode shown across the top row (marked (a), Episode 29 of the 50 in the Goal task suite).
Here, confidence begins high, but falls until it is slightly below the halting threshold at 20\% task completion, coinciding with the difficult subtask of gripping the rounded glass bottle.
Allowed to continue, the model is able to grip the bottle, and confidence rebounds as it approaches the wine rack.
However, confidence once again falls quickly when it begins to set the bottle down, possibly because the difficult grip put the bottle in an unfavorable position for placement.
In the end, the robot fails to place the bottle securely on the rack, with potentially negative real-world consequences.
Given the opportunity, a safety intervention could have been performed at multiple points before this incident, either to avoid grasping the bottle in the first place or to reset the bottle safely on the table instead of trying a failed placement.

\clearpage

\subsection{Calibration Across Action Dimensions}

\begin{figure*}[t]
\centering
\includegraphics[width=0.7\textwidth]{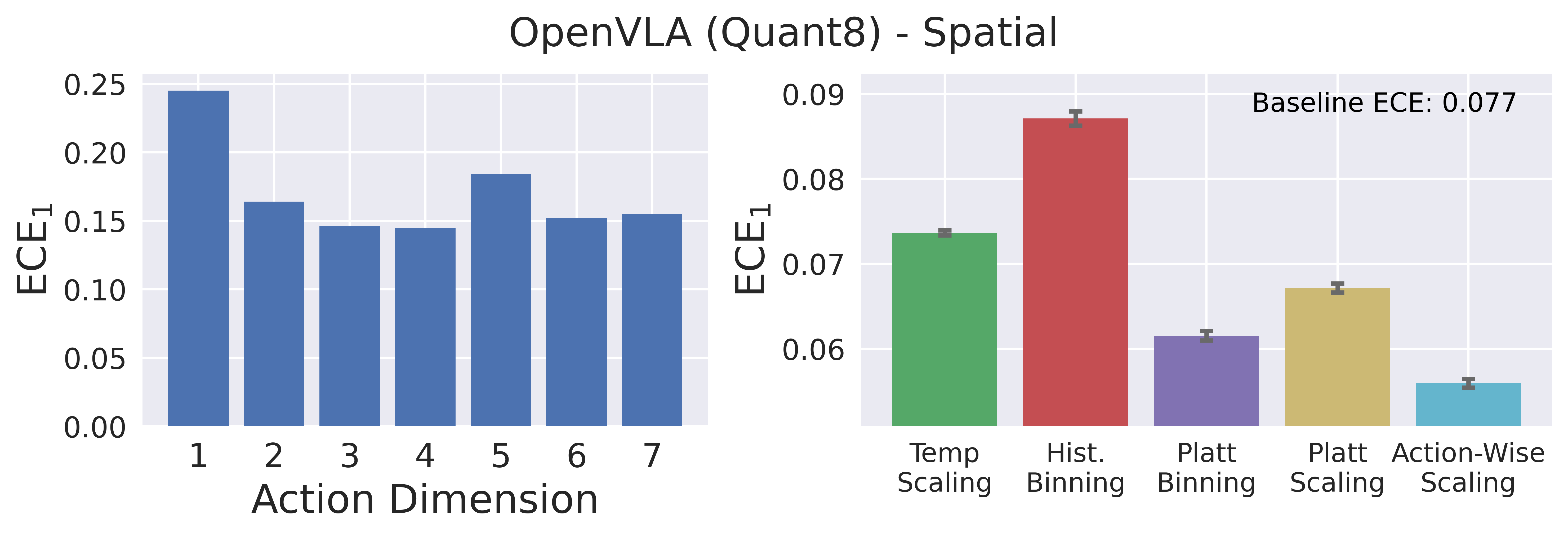}
\includegraphics[width=0.7\textwidth]{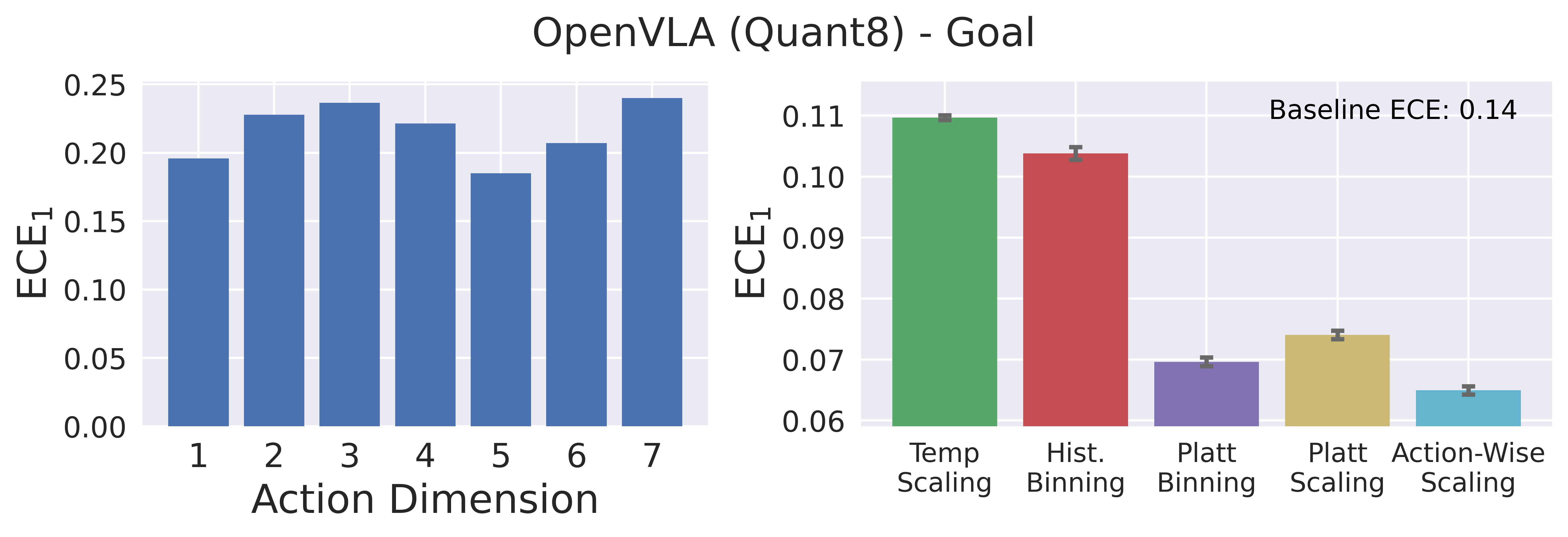}
\includegraphics[width=0.7\textwidth]{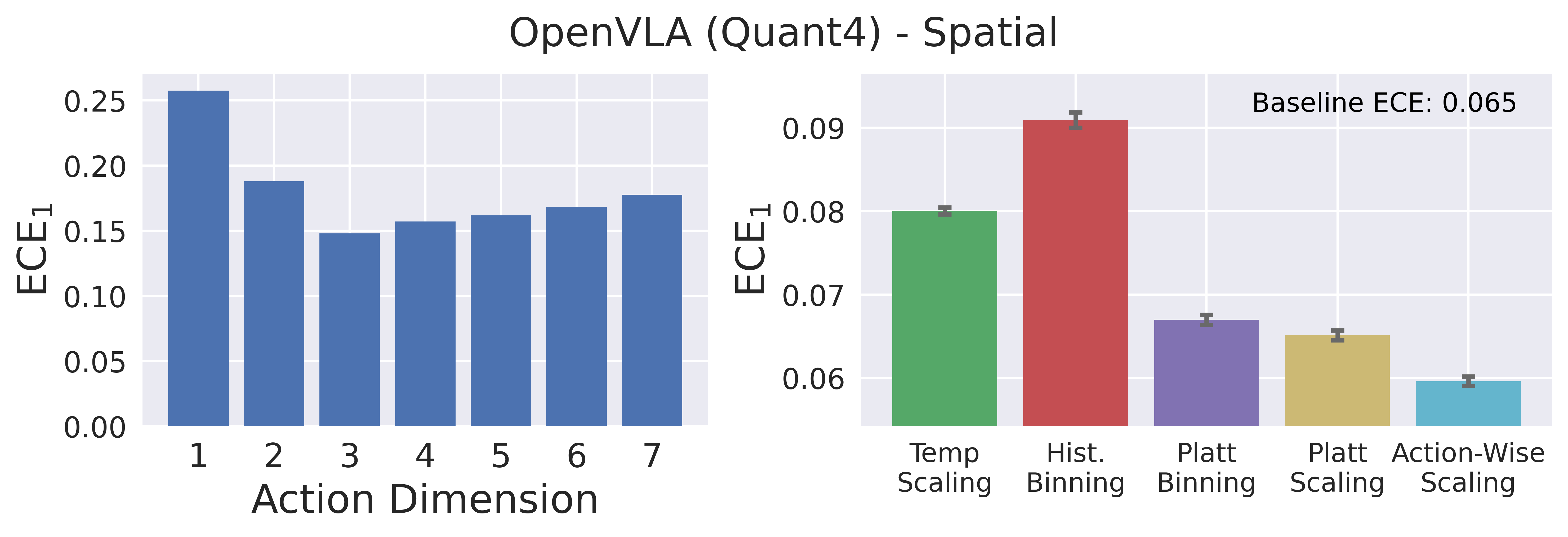}
\includegraphics[width=0.7\textwidth]{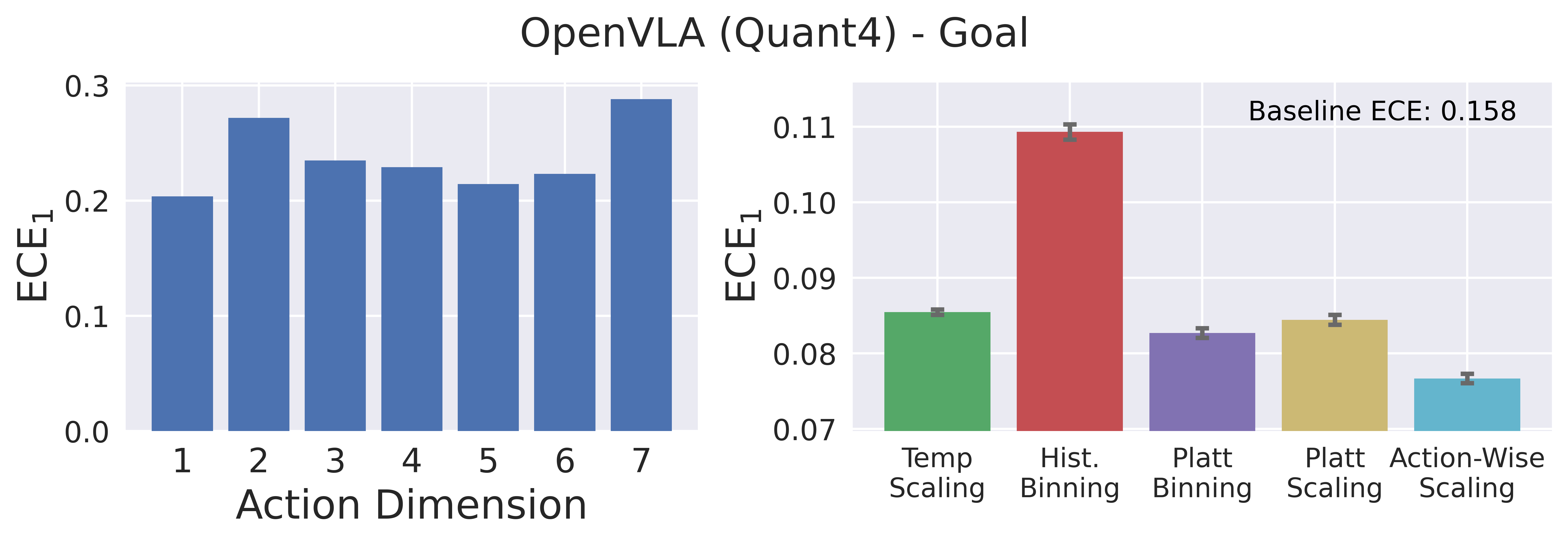}
\caption{On the left of each pair of plots, we compare miscalibration across action dimensions.  On the right, we compare the performance of typical Platt scaling, temperature scaling, histogram binning, and Platt binning to action-wise Platt scaling.}
\label{fig:scaling_appendix}
\end{figure*}

Figure~\ref{fig:scaling_appendix} shows additional results for the action-wise recalibration experiments.

\end{document}